\documentclass[sigconf]{acmart}

\usepackage{booktabs} 

\usepackage{graphicx}
\usepackage{subcaption}
\usepackage{amssymb,amsmath,latexsym,amsthm}
\usepackage{comment}

\newcommand{\etal}{\textit{et al. }}

\newcommand{\heading}[1]{{\medskip \noindent \textbf{#1}}}
\newcommand{\reminder}[1]{{\textcolor{blue}{#1}}}

\newcommand{\squishlist}{
   \begin{list}{$\bullet$}
    { \setlength{\itemsep}{0pt}      \setlength{\parsep}{3pt}
      \setlength{\topsep}{3pt}       \setlength{\partopsep}{0pt}
      \setlength{\leftmargin}{1.0em} \setlength{\labelwidth}{1em}
      \setlength{\labelsep}{0.5em} } }
\newcommand{\squishend}{
    \end{list}  }

\newcommand{\squisholist}{
        \begin{enumerate}
                { \setlength{\itemsep}{0pt}      \setlength{\parsep}{0pt}
                        \setlength{\topsep}{0pt}       \setlength{\partopsep}{0pt}
                        \setlength{\leftmargin}{0em} \setlength{\labelwidth}{0em}
                        \setlength{\labelsep}{0em} } }
        \newcommand{\squishendo}{
        \end{enumerate}  }

\acmYear{}
\copyrightyear{}
\setcopyright{none}
\acmConference{}{}{}
\acmPrice{}
\acmDOI{}
\acmISBN{}

\begin{document}
\title{Adversarial Perturbations Against Real-Time Video Classification Systems}

\author{Shasha Li}
\affiliation{%
\institution{University of California, Riverside}
  \city{Riverside}
  \state{California}
}
\email{sli057@ucr.edu}

\author{Ajaya Neupane}
\affiliation{%
  \institution{University of California, Riverside}
  \city{Riverside}
  \state{California}
}
\email{ajaya@ucr.edu}

\author{Sujoy Paul}
\affiliation{%
  \institution{University of California, Riverside}
  \city{Riverside}
  \state{California}
}
\email{spaul003@ucr.edu}

\author{Chengyu Song}
\affiliation{%
  \institution{University of California, Riverside}
  \city{Riverside}
  \state{California}
}
\email{csong@cs.ucr.edu}

\author{Srikanth V. Krishnamurthy}
\affiliation{%
  \institution{University of California, Riverside}
  \city{Riverside}
  \state{California}
}
\email{krish@cs.ucr.edu}

\author{Amit K. Roy Chowdhury}
\affiliation{%
  \institution{University of California, Riverside}
  \city{Riverside}
  \state{California}
}
\email{amitrc@ece.ucr.edu}

\author{Ananthram Swami}
\affiliation{%
  \institution{United States Army Research Laboratory}
}
\email{ananthram.swami.civ@mail.mil}
\renewcommand{\shortauthors}{S. Li et al.}

\begin{abstract}
{
Recent research has demonstrated the brittleness of machine learning systems
to adversarial perturbations. However, the studies have been mostly limited to
perturbations on images and more generally, classification that does not deal
with temporally varying inputs. In this paper we ask "Are adversarial perturbations
possible in real-time video classification systems and if so, what properties must
they satisfy?" Such systems find application in surveillance applications, smart vehicles,
and smart elderly care 
and thus,
misclassification could be particularly harmful (e.g., a mishap at an elderly
care facility may be missed). We show that accounting for temporal
structure is key to generating adversarial examples in such systems. We exploit
recent advances in generative adversarial network (GAN) architectures to account for
temporal correlations and generate adversarial samples that can cause misclassification
rates of over 80 \% for targeted activities. More importantly, the samples also leave
other activities largely unaffected making them extremely stealthy. Finally, we also surprisingly
find that in many scenarios, the same perturbation can be applied to every frame in a video clip that makes
the adversary's ability to achieve misclassification relatively easy.
}
\end{abstract}

\maketitle

\section{Introduction}



Deep Neural Networks (DNN) have found an increasing role in 
real world applications for the purposes of real-time video classification.
Examples of such applications include video surveillance~\cite{sultani2018real},
self driving cars \cite{kataoka2018temporal}, health-care~\cite{umboocv2016case}, etc.
To elaborate, video surveillance systems capable of automated detection of undesired human
behaviors (e.g., violence) can trigger alarms and drastically reduce information workloads on human operators.
Without the assistance of DNN-based classifiers,
human operators will need to simultaneously view and assess footage from a large number of video sensors.
This can be a difficult and exhausting task,
and comes with the risk of missing behaviors of interest and slowing down decision cycles.
In self-driving cars, {video classification} has been used to
understand pedestrian actions and make navigation decisions \cite{kataoka2018temporal}.
Real-time video classification systems have also been deployed for automatic
``fall detection'' in elderly care facilities~\cite{umboocv2016case},
and abnormal detection around automated teller machines~\cite{tripathi2016real}.
All of these applications directly relate to the physical security or safety of people and property.
Thus, stealthy attacks on such real-time {video classification} systems are
likely to cause unnoticed pecuniary loss and compromise personal safety.


Recent studies have shown that virtually all DNN-based systems are vulnerable to
well-designed adversarial inputs~\cite{moosavi2017universal,moosavi2016deepfool,
sharif2016accessorize, goodfellow2014generative,szegedy2013intriguing},
which are also referred to as \emph{adversarial examples}.
Szegedy \etal \cite{szegedy2013intriguing} showed that adversarial perturbations
that are hardly perceptible to humans can cause misclassification in DNN-based image classifiers.
Goodfellow \etal \cite{goodfellow2014explaining} analyzed the potency of
realizing adversarial samples in the physical world.
Moosavi \etal \cite{moosavi2017universal} and Mopuri \etal \cite{mopuri2017nag} introduced
the concept of ``image-agnostic'' perturbations.
As such, the high level question that we try to answer in this paper is
``Is it possible to launch stealthy attacks against DNN-based real-time
video classification systems, and if so how?''


Attacking a video classifier is more complex than attacking an image classifier,
because of the presence of the temporal dimension in addition to
the spatial dimensions present in 2D images.
Specifically, attacking a real-time video classifier poses additional challenges.
\textit{First}, because the classification is performed in real-time,
the corresponding perturbations also need to be generated on-the-fly
with the same frame rate which is extremely computationally intensive.
\textit{Second}, to make the attack stealthy,
attackers would want to add perturbations on the video {in} such a way that
they will only cause misclassification for the targeted (possibly malicious) actions,
while keep the classification of other actions unaffected.
In a real-time video stream, since the activities change across time,
it is hard to identify online and in one-shot \cite{fanello2013one},
the target frames on which to add perturbations.
\textit{Third}, video classifiers use video clips (a set of frames)
as inputs \cite{fanello2013one,tripathi2016real}.
As video is captured, it is broken up
into clips and each clip is fed to the classifier.
As a result, even if attackers are aware of the length of each clip
(a hyper-parameter of the classifier),
it is hard to predict {\em when} each clip begins and ends.
Therefore, if they generate perturbations for a clip using traditional
methods (e.g., gradient descent), the perturbations might not work because
they are not aligned with the real clip the classifier is using
(Please see~\autoref{fig:sliding_window} and the associated discussion for more details).

In this paper, our first objective is to investigate how to generate
adversarial perturbations against real-time video classification
systems by overcoming the above challenges.
We resolve the first (real-time) challenge by using {\em universal perturbations}~\cite{moosavi2017universal}.
Universal perturbations allow us to affect the classification results
using a (single) set of perturbations generated off-line.
Because they work on unseen inputs they preclude the need for intensive on-line
computations to generate perturbations for every incoming video clip.
To generate such universal perturbations, we leverage the generative
adversarial network (GAN) \cite{goodfellow2014generative} architecture.


However, adding universal perturbations on all the frames of the video can cause
the misclassification of all the actions in the video stream.
This may expose the attack as the results may not make sense (e.g., many people performing rare actions).
To make the attack stealthy,
we introduce the novel concept of \textit{dual purpose universal perturbations},
which we define as universal perturbations which only cause the misclassification for
inputs belonging to the target class,
while minimize, or ideally, have no effect on the classification results
for inputs belonging to the other classes.

Dual purpose perturbations by themselves do not provide high success rates
in terms of misclassification because of challenge three, which is that the mis-alignment of
the boundaries of perturbations with respect to the real clip boundaries (input to the classifier)
significantly affects the misclassification success rates.
To solve this problem, we introduce a new type of perturbation that
we call the \textit{Circular Universal Dual Purpose Perturbations (C-DUP)}.
The C-DUP is a 3D perturbation (i.e., a perturbation clip composed of a sequence of frames),
which is a valid perturbation on a video regardless of {the start and end of each clip}.
In other words, it works on all cyclic permutations of frames in a clip.
To generate the C-DUP, we make significant changes to the baseline GAN architecture.
In particular, we add a new unit to generate circular perturbations,
that is placed between the generator and the discriminator (as discussed later).
We demonstrate that the C-DUP {\color{black} is very {stable and effective}
in {achieving real-time stealthy attacks on video classification systems}.

After demonstrating the feasibility of stealthy attacks against real-time
video classification systems, our second objective is to investigate the effect
of the temporal dimension.
In particular, we investigate the feasibility of attacking the classification
systems using a simple and light 2D perturbation
which is applied across all the frames of a video.
By tweaking our generative model, we are able to generate such perturbations
which we name as \textit{2D Dual Purpose Universal Perturbations (2D-DUP)}.
These perturbations work well on a sub-set of videos, but not all.
We will discuss the reasons for these 2D attacks in \S~\ref{sec:2D}.

\medskip
\noindent \textbf{Our Contributions:}
In brief, our contributions in this paper are:

\squishlist
	\item We provide a comprehensive analysis on the challenges in crafting
	adversarial perturbations for real-time video classifiers.
	We empirically identify what we call the boundary effect phenomenon in
	generating adversarial perturbations against video (see~\S~\ref{sec:temporal}).

	\item We design and develop a generative framework to craft two types of
	stealthy adversarial perturbations against real-time video classifiers, viz.,
	circular dual purpose universal perturbation (C-DUP)
	and 2D dual purpose universal perturbation (2D-DUP).
	These perturbations are agnostic to (a) the video captured (universal) and
	(b) the temporal sequence of frames in the clips input to the video classification system
	(resistance to cyclic permutations of frames in a clip).

	\item We demonstrate the potency of our adversarial perturbations using two different video datasets.
	In particular, the UCF101 dataset captures coarse-grained activities
	(human actions such as applying eye makeup, bowling, drumming) \cite{soomro2012ucf101}.
	The Jester dataset captures fine-grained activities
	(hand gestures such as sliding hand left, sliding hand right, turning hand clockwise, turning hand counterclockwise) \cite{jester2016}.
	We are able to launch stealthy attacks on both datasets 
	with over a 80 \% misclassification rate,
	while ensuring that the other classes are correctly classified with {relatively high accuracy}.
\squishend

\section{Background}

In this section, we provide the background relevant to our work. Specifically, we discuss how a real-time video classification system works and what standard algorithms are currently employed for action recognition. We also discuss generative adversarial networks (GANs) in brief. 

\subsection{Real-time video-based classification systems}
\label{sec:var}

DNN based video classification systems are being increasingly deployed in real-world scenarios. 
Examples include fall detection in elderly care \cite{foroughi2008intelligent}, abnormal event 
detection on campuses \cite{umboocv2017caseNCHU,umboocv2017caseTai}, 
security surveillance for smart cities \cite{umboocv2017caseCBS}, and self-driving cars \cite{kataoka2018temporal,kataoka2018drive}. 
Given an input real-time video stream, which may contain one or more known actions,
the goal of a video classification system is to correctly recognize the sequence of the
performed actions. Real-time video classification systems commonly use a {\em sliding window} 
to analyze a video stream \cite{fanello2013one,tripathi2016real}. 
The classifier computes an output score for each class in each
sliding window. 
The sliding window moves with a stride. 
Moving in concert with the sliding window, one can generate ``score curves'' for each 
action class. Note that the scores for all the action classes evolve with time. The score curves are then smoothed (to remove noise) as shown in \autoref{fig:real_time}. With the smoothed score curves, 
the on-going actions are predicted online.
From the figure one can see that, the real-time video classification system is fooled if one can make the classifier output a low score for the true class in each sliding window; with this, the true actions will not 
be recognized.
\begin{figure}[!ht]
    \centering
    \includegraphics[scale=.36]{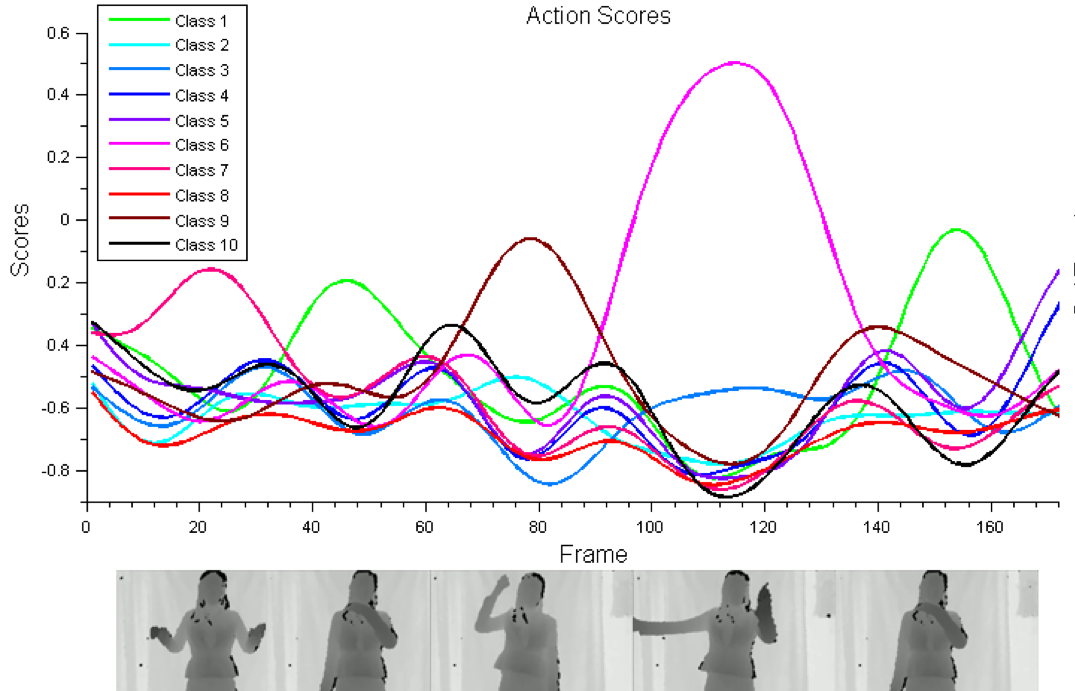}
    \caption{This figure \cite{fanello2013one} illustrates the score curves computed by a video classifier with a sliding window for every class. Real-time video classification systems use these score curves to do online action recognition. }
    \label{fig:real_time}
\end{figure}



\begin{figure*}[!ht]
	\centering
	\includegraphics[scale=.36]{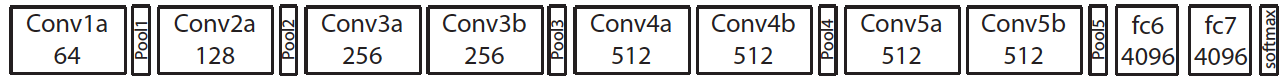}
	\caption{The C3D architecture\cite{tran2015learning}.
	C3D net has 8 convolution, 5 max-pooling, and 2 fully connected layers, followed by a softmax output layer. All 3D convolution kernels are $3 \times 3 \times 3$ with a stride \cite{tran2015learning} of 1 in both spatial and temporal dimensions. The number of filters are shown in each box. The 3D pooling layers are represented from pool1 to pool5. All pooling kernels are $2 \times 2 \times 2$, except for pool1, which is $1 \times 2 \times 2$. Each fully connected layer has 4096 output units.}
	\label{fig:c3d}
\end{figure*}

\subsection{The C3D classifier}\label{sec:c3d} 

Next, 
we describe what is called the C3D classifier~\cite{tran2015learning}, a state of the art
classifier that we target in our paper. 

DNNs and in particular convolutional neural networks (CNNs) are being increasingly applied in video 
classification.
Among these, spatio-temporal networks like C3D \cite{tran2015learning} and two-stream networks like I3D \cite{carreira2017quo} outperform other network structures\cite{herath2017going,gu2017recent}. 
Without the requirement of non-trivial pre-processing on the video stream, 
spatio-temporal networks demonstrate high efficiency; among these, C3D is the start-of-art model~\cite{gu2017recent}.
 
The
C3D model is generic, which means that it can differentiate across different types of videos (e.g., videos of actions in sports, actions involving pets, actions relating to food etc.). 
It also provides a compact representation that facilitates scalability in processing, storage, and retrieval. It is also extremely efficient in classifying video streams
(needed in real-time systems).

Given its desirable attributes and popularity, without loss of generality, 
we use the C3D model as our attack target in this paper.
The C3D model is based on 3D ConvNet (a 3D CNN) \cite{karpathy2014large,tran2015learning,varol2017long}, which is very effective in modeling 
temporal information 
(because it employs 3D convolution and 3D pooling operations). 

The architecture and hyperparamters of C3D are shown in \autoref{fig:c3d}. The input to the C3D 
classifier is a clip consisting of 16 consecutive frames. This means that upon using C3D, 
the sliding window size is 16. Both the height and the width of each frame are 112 
pixels and each frame has 3 (RGB) channels. 
The last layer of C3D is a softmax layer that provides a classification score 
with respect to each class.

\subsection{Generative Adversarial Networks and their relevance}
\label{sec:nag}

Generative Adversarial Networks (GANs) \cite{goodfellow2014generative} 
were initially developed to generate synthetic content (and in particular, images) that conformed
with the space of natural content~\cite{zhu2016generative,huang2017stacked}. Recently,
there has also been work on using GANs for generating synthetic videos~\cite{vondrick2016generating}.
The GAN consists of two components viz., a generator or generative model that tries to learn
the training data distribution and a discriminator or discriminative model that seeks to 
distinguish the generated distribution from the training data distribution. 

To elaborate, consider the use of a GAN for video generation. An archtecture for this purpose is
shown in \autoref{fig:gan_arc}. The generator $G$, learns a map 
from a random vector $z$ in latent space, to a natural video clip $V$; in other words, $G(z)=V$, where $z$ is usually sampled from a simple distribution such as a Gaussian distribution ($N(0,1)$)
or a uniform distribution ($U(-1,1)$). The discriminator $D$ takes a video clip as input
(either generated or natural),
and outputs the probability that it is from the training data (i.e., the probability that
the video clip is natural).
The interactions between $G$ and $D$ are modeled as a game and in theory, at the end $G$ must be able to generate video clips from the true training set distribution. 

Recently, a GAN-like architecture has been used to generate adversarial perturbations on images \cite{mopuri2017nag} .
The authors of ~\cite{mopuri2017nag} keep a trained discriminator fixed; 
its goal is to classify the inputs affected by the perturbations from the
generator. The generator learns from the discriminator classifications to modulate its perturbations.
Similarly,  we apply the GAN-like architecture wherein
the objective is to allow the generator to learn a distribution for candidate perturbations that can
fool our discriminator. Our model incorporates significant extensions to the GAN structure
used in \cite{mopuri2017nag} to account for the unique properties of video classification that
were discussed earlier.

\section{Threat Model and Datasets}

In this section, we describe our threat model. We also provide a brief overview of the datasets we chose for validating our attack models.

\subsection{Threat model}

We consider a white-box model for our attack, i.e.,
the adversary has access to the training datasets used to train the video classification system,
and has knowledge of the deep neural network model the real-time classification system uses.
We also assume that the adversary is capable of injecting perturbations in the real-time video stream.
In particular, we assume the adversary to be a man-in-the-middle that can
intercept and add perturbations to streaming video \cite{mitm2014videos},
or that it could have previously installed a malware that is able to add perturbation
prior to classification \cite{malware2016zdnet}.

We assume that the goal of the adversaries is to launch stealthily attacks,
i.e., they want the system to only misclassify the malicious actions
without affecting the recognition of the other actions.
So, we consider two attack goals.
\textit{First}, given a target class, we want all the clips from this class
to be misclassified by the real-time video classifier.
\textit{Second}, for all the clips from other (non-target) classes,
we want the classifier to correctly classify them.



\subsection{Our datasets} 
\label{sec:dataset}
We use the human action recognition dataset UCF-101 \cite{soomro2012ucf101} and the hand gesture recognition dataset 20BN-JESTER dataset (Jester) \cite{jester2016} to validate our attacks on 
video classification systems. We use these two datasets because they represent two kinds of classification, i.e., coarse-gained and fine-grained action classification. 

\heading{The UCF 101 dataset:}
The UCF 101 dataset used in our experiments is the standard dataset collected from Youtube. It includes 13320 videos from 101 human action categories (e.g., applying lipstick, biking, blow drying hair, 
cutting in the kitchen etc.). The videos collected in this dataset have variations in camera motion, appearance, background, illumination conditions etc. Given the diversity it provides, we consider the dataset to validate the feasibility of our attack model on coarse-gained actions. 
There are three different (pre-existing) splits \cite{soomro2012ucf101} in the dataset; we use split 
1 for both training and testing, in our experiments. The training set includes 9,537 video clips and the testing set includes 3,783 video clips.

\heading{The Jester dataset:}
The 20BN-JESTER dataset (Jester)  is a recently collected dataset with hand gesture videos. These videos are recorded by crowd-source workers performing 27 kinds of gestures (e.g., sliding hand left, sliding two fingers left, zooming in with full hand, zooming out with full hand etc.). We use this dataset to validate our attack with regards to fine-grained actions.
Since this dataset does not currently provide labels for the testing set, 
we use the validation set as our testing set (i.e., apply our perturbations on the validation set). 
The training set has 148,092 short video clips and our testing set has 14,787 short video clips.


\section{Generating perturbations for real-time video stream}
\label{sec:realtime}
\begin{figure}[t]
	\centering
	\begin{subfigure}{0.5\textwidth}
		\includegraphics[scale=.22]{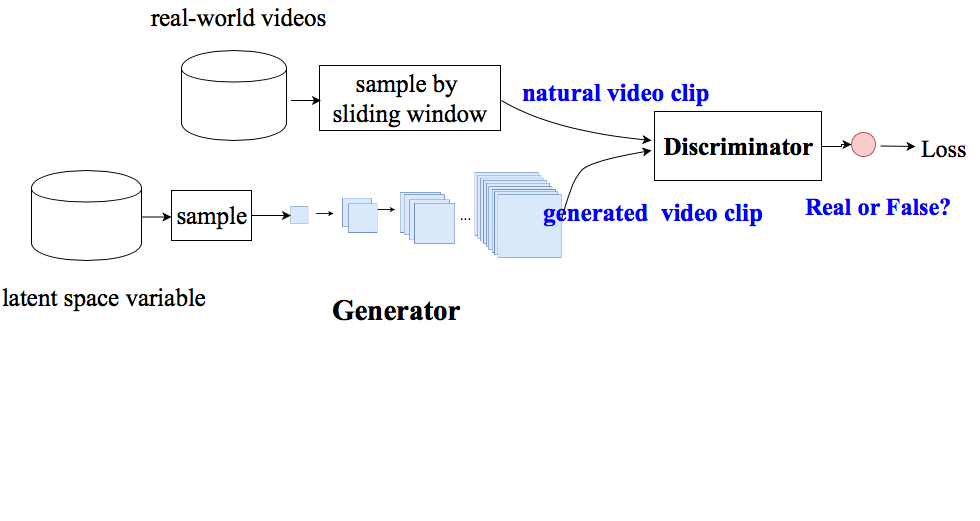}
		\vspace{-15mm}
		\subcaption{GAN Architecture}
		\vspace{5mm}
		\label{fig:gan_arc}    
	\end{subfigure}
	\begin{subfigure}{0.5\textwidth}
		\includegraphics[scale=.20]{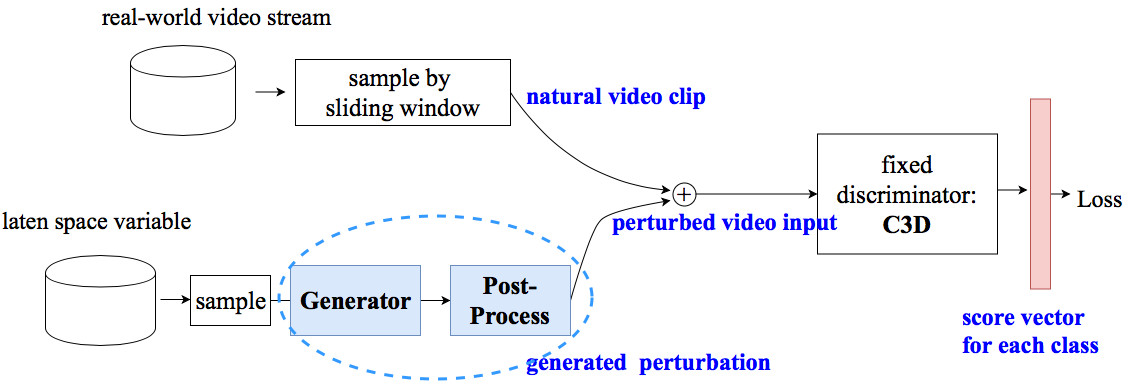}
		\subcaption{Our Architecture}
		\label{fig:our_arc}   
	\end{subfigure}
	\caption{This figure compares our architecture with traditional GAN architectures. Our architecture is different fromtraditional GAN in the following aspects: 1) The discriminator is a pre-trained classifier we attack, whose goal is to classify videos, and not to distinguish between the natural and synthesized inputs; 2) The generator generates perturbations, and not direct inputs to the discriminator, and the perturbed training inputs are fed to discriminator; 3) The learning objective is to let the discriminator misclassify the perturbed inputs.}
	\label{fig:arc}
\end{figure}

From the adversary's perspective, we first consider
the challenge of attacking a real-time video stream.
In brief, when attacking an image classification system, the attackers usually take the 
following approach. First, they obtain the {\em target} image that is to be attacked with its true label. 
Next, they formulate a problem wherein they try to compute the ``minimum'' noise that is to be
added in order to cause a mis-classification of the target.
The formulation takes into account the function of the classifier, the input image, and its true label. 
In contrast, in the context of real-time video classification, the video is not available to the attackers a priori.
Thus, they will need to create perturbations that can effectively perturb an incoming video stream,
whenever a {\em target class} is present.

Our approach is to compute the perturbations offline and apply them online.
Since we cannot predict what is captured in the video, we need perturbations
which work with unseen inputs. A type of perturbations that satisfies this requirement
is called the Universal Perturbation (UP), which has been studied in
the context of generating adversarial samples against image classification
systems~\cite{mopuri2017nag,moosavi2017universal}.
In particular, Mopuri \etal, 
have developed a generative model that learns the space of universal perturbations for images
using a GAN-like architecture.

Inspired by this work, we develop a similar architecture,
but make modifications to suit our objective.
Our goal is to generate adversarial perturbations that fool the
discriminator instead of exploring the space for diverse UPs.
In addition, we retrofit the architecture to handle video inputs.
Our architecture is depicted in \autoref{fig:our_arc}.
It consists of three main components:
1) a 3D generator which generates universal perturbations;
2) a post-processor, which for now does not do anything but is needed
to solve other challenges described in subsequent sections; and
3) a pre-trained discriminator for video classification,
e.g., the C3D model described in \S~\ref{sec:c3d}.
Note that unlike in traditional GANs
wherein the generator and the discriminator are trained together,
we only train the generator to generate universal perturbations to
fool a fixed type of discriminator.

The 3D generator in our model 
is configured to use 3D deconvolution layers and provide 3D outputs as shown in \autoref{fig:arc_roll}.
Specifically, it generates a clip of perturbations,
whose size is equal to the size of the video clips taken as input by the C3D classifier.
To generate universal perturbations, the generator first takes a noise vector $z$ from a latent space.
Next, It maps $z$ to a perturbation clip $p$, such that, $G(z)=p$.
It then adds the perturbations on a training clip $x$ to obtain the perturbed clip $x+p$.
Let $c(x)$ be the true label of $x$.
This perturbed clip is then input to the
C3D model which outputs the score vector $Q(x+p)$ (for the perturbed clip).
The classification should ensure that the highest score corresponds to
the true class
($c(x)$ for input $x$) 
in the benign setting.
Thus, the attacker seeks to generate a $p$ such that the C3D classifier outputs
a low score to the $c(x)$th element in $Q$ vector (denoted as $Q_{c(x)}$) for $x+p$.
In other words, this means that after applying the perturbation,
the probability of mapping $x$ to class $c(x)$ is lower than the
probability that it is mapped to a different class
(i.e., the input is not correctly recognized).

We seek to make this perturbation clip $p$ ``a universal perturbation'',
i.e., adding $p$ to any input clip belonging to the target class
would cause misclassification.
This means that we seek to minimize the sum of the cross-entropy loss
over all the training data as per \autoref{equ:cross_entr}.
Note that the lower the cross-entropy loss,
the higher the divergence of the predicted probability from the true label~\cite{huang2017stacked}.

\begin{equation}
\underset{G}{\text{minimize}}\quad \sum_{x\in X}{-\log[1-Q_{c(x)}(x+G(z)]}
\label{equ:cross_entr}
\end{equation}

When the generator is being trained, for each training sample,
it obtains feedback from the discriminator and adjusts its parameters 
to cause the discriminator to misclassify that sample.
It tries to find a perturbation that works for every sample
from the distribution space known to the discriminator.
At the end of this phase, the attacker will have a generator
that outputs universal perturbations which can cause the misclassification on any
incoming input sample from the same distribution (as that of the training set).
However, as discussed next, just applying the universal perturbations alone
will not be sufficient to carry out a successful attack.
In particular, the attack can cause unintended clips to be misclassified as well,
which could compromise our stealth requirement as discussed next in \S \ref{sec:perturbframe}.

\section{making perturbations stealthy}
\label{sec:perturbframe}

Blindly adding universal perturbations will affect the classification of
clips belonging to other non-targeted classes. 
This may raise alarms, especially if many of such misclassifications are mapped on to rare actions.
Thus, while causing the target class to be misclassified,
the impact on the other classes must be imperceptible.
This problem can be easily solved when dealing
with image recognition systems since images are self-contained entities,
i.e., perturbations can be selectively added on target images only.
However, video inputs change temporally and an action captured in a set of
composite frames may differ from that in the subsequent frames.
It is thus hard to a priori identify (choose) the frames relating to the target class,
and add perturbations specifically on them.
For example, consider a case with surveillance in a grocery store.
If attackers seek to misclassify an action related to shoplifting and
cause this action to go undetected,
they do not have a priori knowledge about when this action will be captured.
And adding universal perturbations blindly could cause mis-classifications of other actions (e.g.,
other benign customer actions may be mapped onto shoplifting actions thus triggering alarms).

Since it is hard (or even impossible) to a priori identify the frame(s)
that capture the intended actions and choose them for perturbation,
the attackers need to add perturbations on each frame.
However, to make these perturbations furtive,
they need to ensure that the perturbations added only mis-classifies
the target class while causing other (non-targeted) classes to be classified correctly.
We name this unique kind of universal perturbations as  ``Dual-Purpose Universal Perturbations''
or DUP for short.

In order to realize DUPs, we have to guarantee that for the input clip $x_t$,
if it belongs to the target class (denote the set of inputs from target class as $T$),
the C3D classifier returns a low score with respect to the correct class $c(x_t)$,
i.e., $Q_{c(x_t)}$.
For input clips $x_s$ that belongs to other (non-target) classes
(denote the set of inputs from non-target classes as $S$, thus, $S=X-T$),
the model returns high scores with regards to their correct mappings ($Q_{c(x_s)}$).
To cause the generator to output DUPs, we refine the optimization problem in \autoref{equ:cross_entr}
as shown in \autoref{equ:dual}:

	\begin{equation}
	\begin{aligned}
	\underset{G}{\text{minimize}} \quad & \lambda\times \sum_{x_t\in T}{-\log[1-Q_{c(x_t)}(x_t+G(z))]}\\
	                              &+\sum_{x_s\in S}{-\log[Q_{c(x_s)}(x_s+G(z))]}
	\end{aligned}
	\label{equ:dual}
	\end{equation}


The first term in the equation again relates to minimizing the cross-entropy of the target class,
while the second term maximizes the cross-entropy relating to each of the other classes.
The parameter $\lambda$ is the weight applied with regards to the misclassification of the target class.
For attacks where stealth is more important, we may use a smaller $\lambda$ to guarantee that the
emphasis on the
misclassification probability of the target class is reduced while
the classification of the non-target classes are affected to the least extent possible.


\section{The time machine: handling the temporal dimension}
\label{sec:temporal}

The final challenge that the attackers will need to address in order to effectively generate
adversarial perturbations against real-time video classification systems
is to handle the temporal structure that exists across the frames in a video.
In this section, we first discuss why directly applying existing methods
that target images to generate perturbations against video streams do not work.
Subsequently, we propose a new set of perturbations that overcome this challenge.

\begin{figure}
	\centering
	\begin{subfigure}{0.5\textwidth}
		\includegraphics[scale=.4]{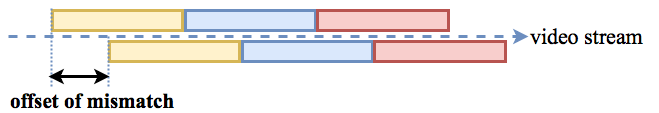}
		\subcaption{This figure depicts a scenario where there is a mismatch between the clip an attacker views to generate a perturbation, and the clip a classifier views for classification.}
		\label{fig:mismatch}
	\end{subfigure}

	\vspace{0.02\textwidth}
	\begin{subfigure}{0.5\textwidth}
		\includegraphics[scale=.4]{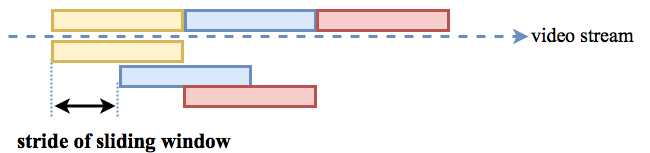}
		\subcaption{This figure displays a scenario where the stride of the sliding window is smaller than the size of a clip. In such a case, the attacker will not know what perturbation to use on a frame, as the frame can be selected in multiple input clips
used in classification.}
		\label{fig:stride_overlap}
	\end{subfigure}
	\caption{The sliding window used to capture clips that are input to the video classification system}
	\label{fig:sliding_window}
\end{figure}
\subsection{The boundary effect
}

The input to the video classifier is a clip of a sequence of frames.
As discussed earlier, given an input clip,
it is possible to generate perturbations for each frame in that clip.
However, the attacker cannot a priori determine the {\em clip boundaries} used
by the video classifier.
In particular, as discussed in \S~\ref{sec:var}, there are
three hyper-parameters that are associated with a sliding window of a video classifier.
They are the {\em window size}, the {\em stride} and the {\em starting position} of the clip.
All three parameters affect the effectiveness of the perturbations.
While the first two parameters are known to a white-box attacker,
the last parameter is an artifact of when the video clip is captured and input to the
classifier and cannot be known a priori.

The perturbation that is to be added to a frame depends on the relative
location of the frame within the clip.
In other words, if the location of the frame changes because of a temporally staggered clip,
the perturbation needs to adjusted accordingly.
Thus at a high level, the boundaries of the clips used by the classifier
will have an effect on the perturbations that need to be generated.
We refer to this phenomenon as the boundary effect.
To formalize the problem, let us suppose that there is a video stream represented by
$\{ ...,f_{i-2}, f_{i-1}, f_i,
f_{i+1},f_{i+2}, ...\}$ where each $f_j$ represents a frame.
The perturbation on $f_i$ that is generated based on
the clip [$f_{i}, f_{i+1}, \cdots, f_{i+w-1}$] to achieve misclassification of
a target action, will be different from the one generated based on the
temporally staggered clip $[f_{i-1}, f_{i}, f_{i+1} \cdots, f_{i+w-2}]$ to
achieve the same purpose.

To exemplify this problem, we consider using the traditional methods
to attack C3D model. In particular, we use the API from the CleverHans
repository \cite{papernot2016cleverhans} to generate video perturbations.
Note that, the perturbations generated by CleverHans are neither universal nor dual-purpose;
they simply generate a specific perturbation given a specific input clip.
We use the basic iteration methods with default parameters.
Our approach is as follows.
We consider all the videos in the UCF-101 testing set.
We consider different boundaries for the clips in the videos
(temporally staggered versions of the clips) and use the Python libraries
from CleverHans to generate perturbations for each staggered version.
Note that the sliding window size for C3D is 16 and thus, there are 16 staggered versions.
We choose a candidate frame, and compute the correlations between the perturbations added in the
different staggered versions. Specifically, the perturbations are tensors and the normalized correlation
between two perturbations is the inner product of the unit-normalized tensors representing the perturbations.

We represent the average normalized correlations in the perturbations (computed across all videos and all frames)
for different offsets (i.e., the difference in the location of the
candidate frame in the two staggered clips) 
in the matrix shown in \autoref{fig:corre}.
The row index and the column index represent the location of the frames
in the two staggered clips.
For example, the entry corresponding to $\{7,7\}$ represents
the case where the frame considered was the $7^{th}$ frame in the two clips.
In this case, clearly the correlation is 1.00.
However, we see that the correlations are much lower if the
positions of the same frame in the two clips are different.
As an example, consider the entry $\{5, 9\}$:
its average normalized correlation is 0.39,
which indicates that the perturbations that CleverHans adds in the
two cases are quite different.

\begin{figure}[!ht]
	\centering
	\includegraphics[scale=0.33]{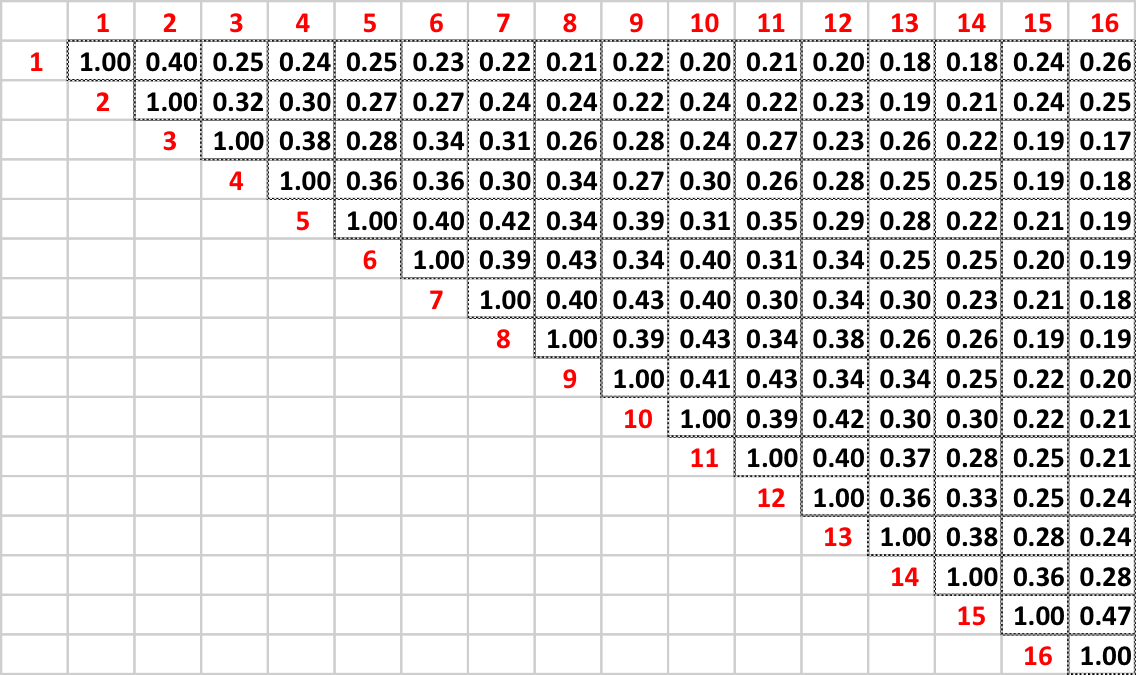}
	\caption{The average normalized correlation matrix computed with perturbations generated using 
the basic iteration API from CleverHans. Both row and column represent the location of frame in clips and the value represents the correlation between perturbations on the same frames but generated when that frame located in different positions in the two temporally staggered clips.}
	\label{fig:corre}
\end{figure}

In \autoref{fig:mag_per}, we show
the average magnitude of perturbations added (over all frames and all videos),
when the target frame is at different locations within a clip.
The abscissa depicts the frame position,
and the ordinate represents the magnitude of the average perturbation.
We observe that the magnitude of the perturbation on the first and last
few frames are larger than those in the middle.
We conjecture this is because the difference in perturbation between
consecutive frame locations are similar
(because the clips with small temporal offsets are similar);
but as the frame location differs by a lot, the perturbation
difference will again go up (clips with larger offsets will be more diverse).

\begin{figure}[!ht]
	\centering
	\includegraphics[scale=0.25]{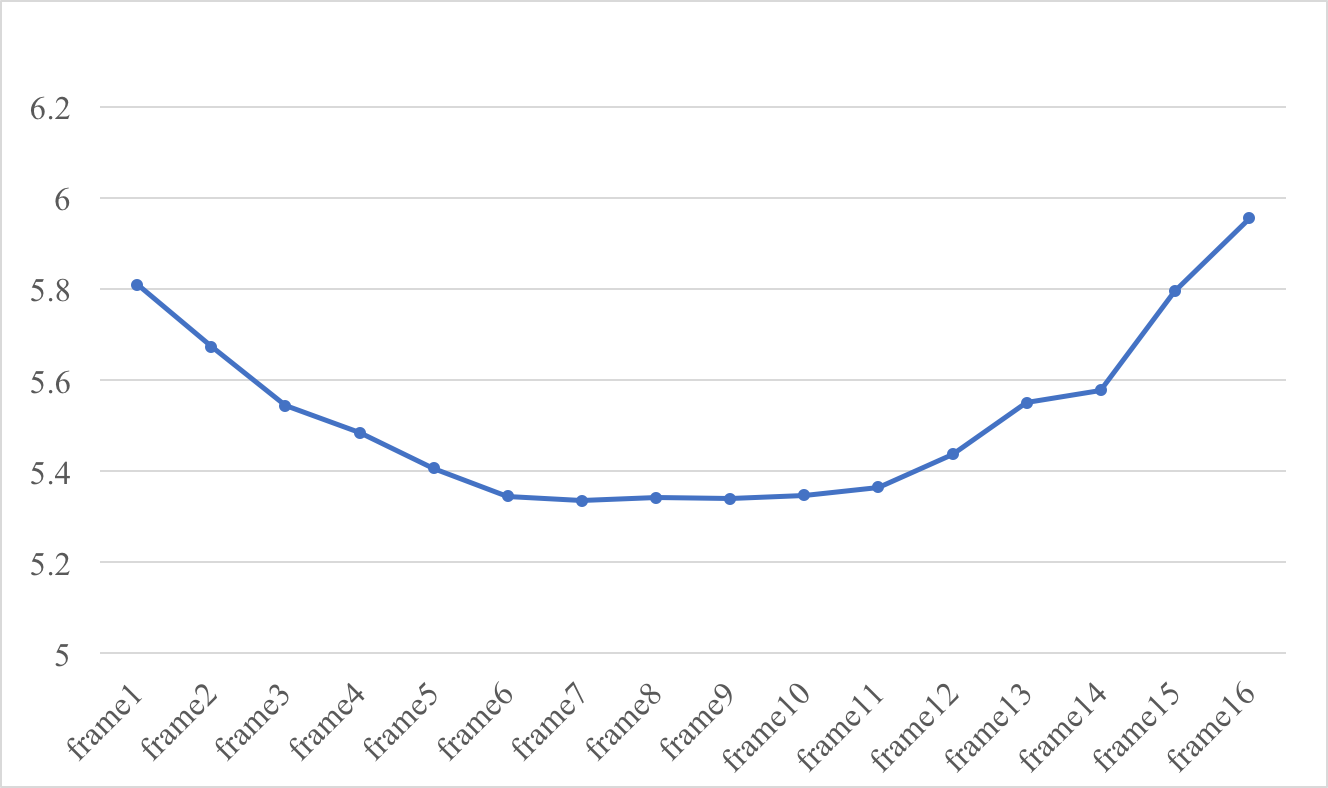}
	\caption{Magnitude of perturbation on each frame: The x axis is the frame position, and the y axis is the magnitude of average perturbation on the frame. }
	\label{fig:mag_per}
\end{figure}

We further showcase the impact of the boundary effect by measuring
the degradation in attack efficacy due to mismatches between
the anticipated start point when the perturbation is generated
and the actual start point when classifying the clip
(as shown in \autoref{fig:mismatch}).
\autoref{fig:dismatch_1} depicts the results.
The abscissa is the offset between the clip considered for
perturbation generation and the clip used in classification.
We can see that as the distance between the two start points increases,
the attack success rate initially degrades but
increases again as the clip now becomes similar to the subsequent clip.
For example, if the offset is 15, the clip chosen 
is almost identical to the subsequent clip.

The boundary effect is also experienced when the stride used by the
classifier is smaller than the clip size.
In brief, stride refers to the extent to which the receiver's sliding
window advances when considering consecutive clips during classification.
If the stride is equal to the clip size,
there is no overlap between the clips that are considered as inputs for classification.
On the other hand, if the stride is smaller than the size of the clip,
which is often the case \cite{tran2015learning,carreira2017quo,fanello2013one,tripathi2016real},
the clips used in classification will overlap with each other
as shown in \autoref{fig:stride_overlap}.
A stride that is smaller than the clip size will cause the same problem
discussed above with respect to the temporal offset between the attackers'
anticipation of the start point and the actual start point of the classifier.
As evident from the figure, such a stride induces an offset,
and this results in the same behaviors that we showcased above.

\begin{figure}[!ht]
	\centering
	\includegraphics[scale=0.3]{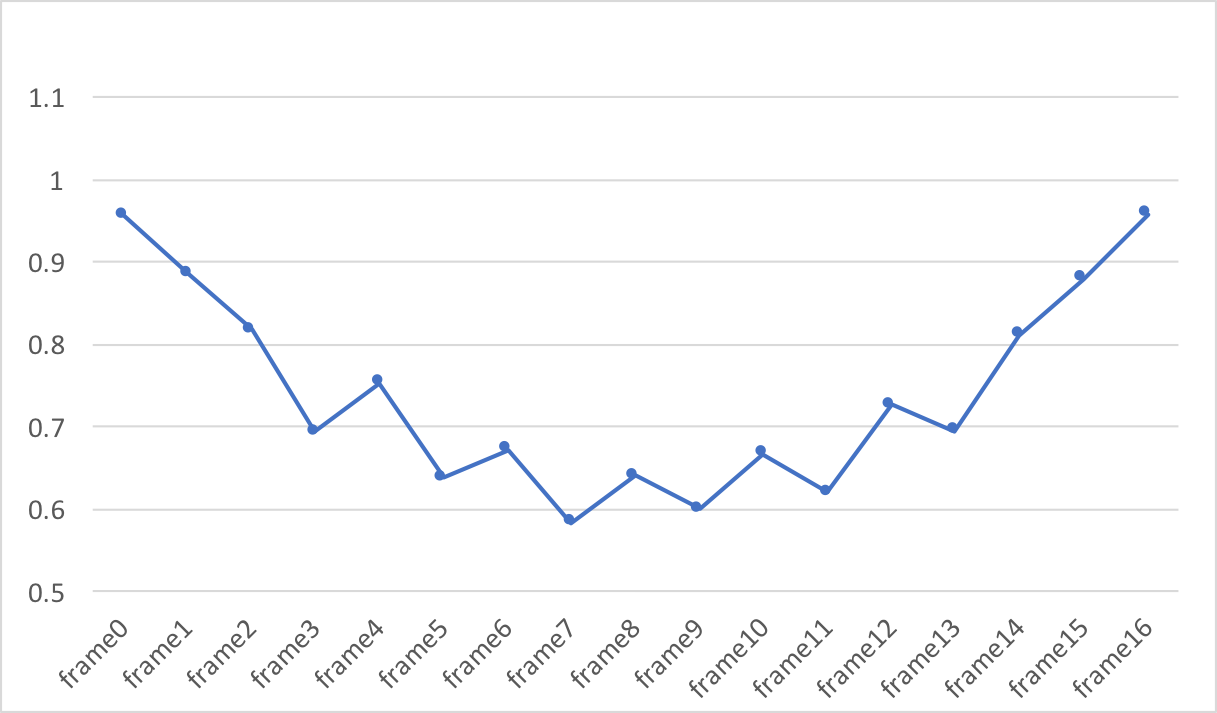}
	\caption{Attack success rate when there is mismatch. The x axis is the offset between the clip generating perturbation and the clip tested. The y axis is the attack success rate.}
	\label{fig:dismatch_1}
\end{figure}

\begin{figure}[ht!]
	\centering
	\includegraphics[scale=.2]{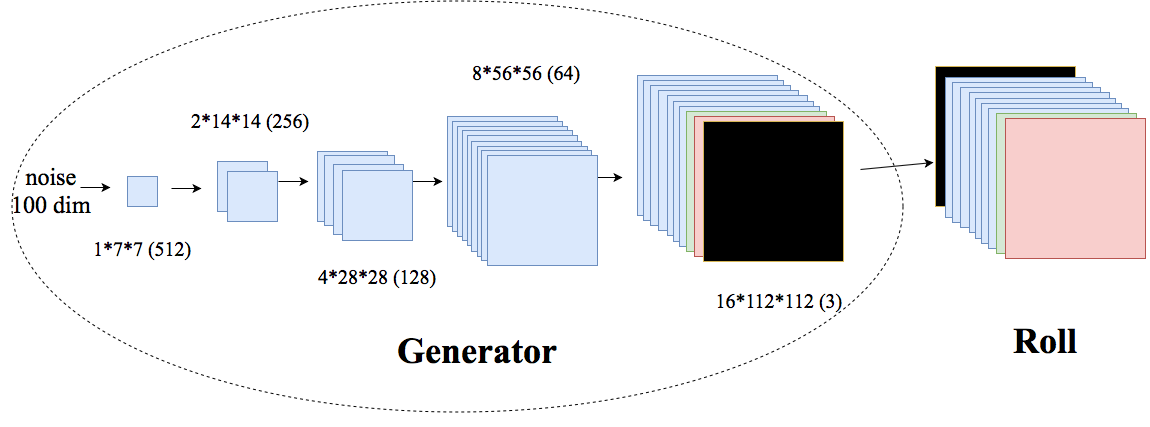}
	\caption{This figure illustrates the Generator and Roll for generating C-DUP.
	1) The generator takes a noise vector as input, and outputs a perturbation clip with 16 frames. The output size for each layer is shown as $\text{temporal dimension} \times \text{horizontal spatial dimension} \times \text{vertical spatial dimension} \times \text{number of channels}$. 2) The roll part shifts the perturbation clip by some offset. The figure shows one example where we roll the front black frame to the back.}
	\label{fig:arc_roll}
\end{figure}
\begin{figure}
	\includegraphics[scale=.22]{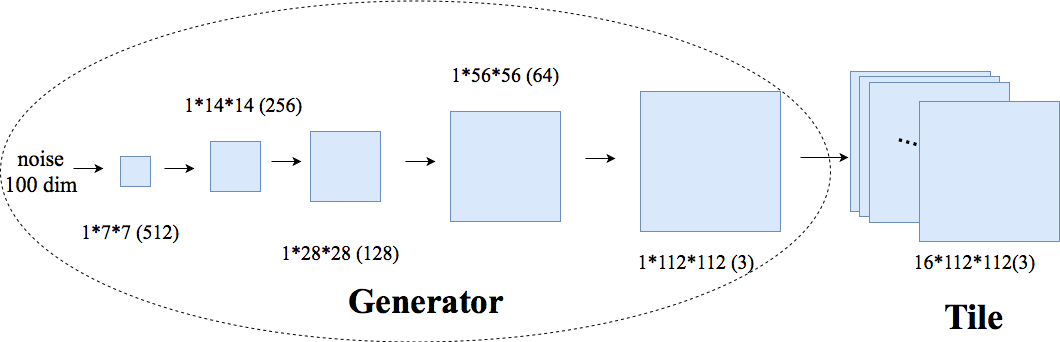}
	\caption{This figure illustrates the Generator and Tile for generating 2D-DUP.
	1) The generator takes a noise vector as input, and outputs a single-frame perturbation. 2) The tile part constructs a perturbation clip by repeating the single-frame perturbation generated 16 times.}
	\label{fig:arc_tile}
\end{figure}

\subsection{Circular Dual-Purpose Universal Perturbation}
\label{sec:CDUP}

The above discussion shows that the boundary effect makes it especially
challenging to attack video classification systems.
To cope with this effect, we significantly extend
the DUPs proposed in \S~\ref{sec:perturbframe} to compose what we call
``Circular Dual-Purpose Universal Perturbations (C-DUP).''

Let us uppose that the size of sliding window is $w$.
Then, the DUP clip $p$ includes $w$ frames (of perturbation),
denoted by $[p_1,p_2, \cdots, p_w]$.
Since $p$ is a clip of universal perturbations,
we launch the attack by repeatedly adding perturbations on each consecutive clip consisting of
$w$ frames, in the video stream.
One can visualize that we are generating a perturbation stream which
can be represented as $[\cdots,p_{w-1}, p_w,p_1,p_2,\cdots ]$.
Now, our goal is to guarantee that the perturbation stream works
regardless of the clip boundaries chosen by the classifier.
In order to achieve this goal,
we need any {\em cyclic or circular shift} of the DUP clip to yield a valid perturbation.
In other words,
we require the perturbation clips $[p_w,p_1,\cdots,p_{w-1}]$, $[p_{w-1},p_w,p_1,\cdots,p_{w-2}]$,
and so on, all to be valid perturbations.
To formalize, we define a permutation function $Roll(p,o)$ which yields
a cyclic shift of the original DUP perturbation by an offset $o$.
In other words, when using $[p_1,p_2, \cdots, p_w]$ as input to $Roll(p,o)$,
the output is $[p_{w-o},p_{w-o+1}, \cdots p_{w}, p{1} \cdots p_{w-o-1}]$.
Now, for all values of $o \in \{0,w-1\}$,
we need $p_o=Roll(p,o)$ to be a valid perturbation clip as well.
Towards achieving this requirement, we use a post-processor unit
which applies the roll function between the generator and the discriminator.
This post processor is captured in the complete architecture is shown in \autoref{fig:our_arc}.

The details of how the generator and the roll unit operate
in conjunction are depicted in \autoref{fig:arc_roll}.
As before, the 3D generator ($G$) takes a noise vector as input and
outputs a sequence of perturbations (as a perturbation clip).
Note that the final layer is followed by a $tanh$ non-linearity which
constrains the perturbation generated to the range [-1,1].
The last thing is to scale the output by $\xi$.
Doing so restricts the perturbation's range to $[-\xi,\xi]$.
Following the work in \cite{moosavi2017universal,mopuri2017nag},
the value of $\xi$ is chosen to be 10 towards making the perturbation quasi-imperceptible.
The roll unit then ``rolls'' (cyclically shifts)
the perturbation $p$ by an offset in $\{0,1,2,\cdots,w-1\}$.
\autoref{fig:arc_roll} depicts the process with an offset equal to 1;
the black frame is rolled to the end of the clip.
By adding the rolled perturbation clip to the training input,
we get the perturbed input.
As discussed earlier, the C3D classifier takes the perturbed input and
outputs a classification score vector.
As before, we want the true class scores to be (a) low for the targeted inputs
and (b) high for other (non-targeted) inputs.
We now modify our optimization function to incorporate the roll function as follows.

\begin{equation}
\begin{aligned}
\underset{G}{\text{minimize}} \quad \sum_{o=1,2\cdots w}\{\lambda\times \sum_{x_t\in T}
{-\log[1-Q_{c(x_t)}(x_t+Roll(G(z),o))]}\\
+\sum_{x_s\in S}{-\log[Q_{c(x_s)}(x_s+Roll(G(z),o))]}\}
\end{aligned}
\end{equation}

The equation is essentially the same as Equation~\ref{equ:dual},
but we consider all possible cyclic shifts of the perturbation output by the generator.

\subsection{2D Dual-Purpose Universal Perturbation}
\label{sec:2D}

We also consider a special case of C-DUP,
wherein we impose an additional constraint which is that ``the perturbations added on
all frames should be the same.''
In other words, we seek to add a single 2D perturbation on each frame
which can be seen as a special case of C-DUP with $p_1 = p_2 = \cdots = p_w$.
We call this kind of DUP as 2D-DUP.
2D-DUP allows us to examine the effect of the temporal dimension in generating
adversarial perturbations on video inputs.

The generator in this case will output a single-frame perturbation
instead of a sequence of perturbation frames as shown in \autoref{fig:arc_tile}.
This is a stronger constraint than the circular constraint,
which may cause the attack success rate to decrease (note that the cyclic property still holds).
However, with this constraint, the perturbation generated is
much simpler and easier to use.

We denote the above 2D perturbation as $p_{2d}$.
The perturbation clip is then generated by simply creating copies of
the perturbation and {\em tiling} them to compose a clip.
The 2D-DUP clip is now $p_{tile} = [p_{2d}, p_{2d}, \cdots, p_{2d}]$
(see \autoref{fig:arc_tile}).
Thus, given that the attack objective is the same as before,
we simply replace the $Roll(p,o)$ function with a $Tile$ function
and our formulation now becomes:

\begin{equation}
\begin{aligned}
\underset{G_{2D}}{\text{minimize}}\quad \lambda\times \sum_{x_t\in T}{-log[1-Q_{c(x_t)}(x_t+Tile(G_{2D}(z)))]}\\
+\sum_{x_s\in S}{-log[Q_{c(x_s)}(x_s+Tile(G_{2D}(z)))]}
\end{aligned}
\end{equation}

\section{Evaluations}
\label{sec:evaluation}
In this section, we showcase the efficacy of the perturbations 
generated by our proposed approaches on both the UCF-101 and Jester datasets.

\begin{figure*}[t]
	\centering
	\begin{subfigure}{0.12\textwidth}
		\includegraphics[scale=.5]{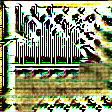}
	\end{subfigure}
	\begin{subfigure}{0.12\textwidth}
		\includegraphics[scale=.5]{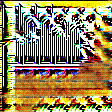}
	\end{subfigure}
	\begin{subfigure}{0.12\textwidth}
		\includegraphics[scale=.5]{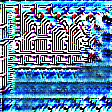}
	\end{subfigure} \vspace{0.01\textwidth}
	\begin{subfigure}{0.12\textwidth}
		\includegraphics[scale=.5]{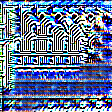}
	\end{subfigure}
	\begin{subfigure}{0.12\textwidth}
		\includegraphics[scale=.5]{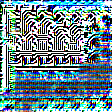}
	\end{subfigure}
	\begin{subfigure}{0.12\textwidth}
		\includegraphics[scale=.5]{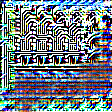}
	\end{subfigure}
	\begin{subfigure}{0.12\textwidth}
		\includegraphics[scale=.5]{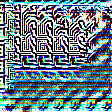}
	\end{subfigure}
	\begin{subfigure}{0.12\textwidth}
		\includegraphics[scale=.5]{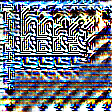} 
	\end{subfigure}
	\caption{DUP on UCF-101}
	\label{fig:v_DUP}
\end{figure*}

\begin{figure*}[t]
	\centering
	\begin{subfigure}{0.12\textwidth}
		\includegraphics[scale=.5]{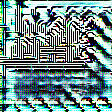}
	\end{subfigure}
	\begin{subfigure}{0.12\textwidth}
		\includegraphics[scale=.5]{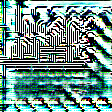}
	\end{subfigure}
	\begin{subfigure}{0.12\textwidth}
		\includegraphics[scale=.5]{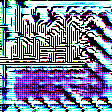}
	\end{subfigure}
	\begin{subfigure}{0.12\textwidth}
		\includegraphics[scale=.5]{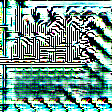}
	\end{subfigure} 	\vspace{0.01\textwidth}
	\begin{subfigure}{0.12\textwidth}
		\includegraphics[scale=.5]{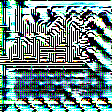}
	\end{subfigure}
	\begin{subfigure}{0.12\textwidth}
		\includegraphics[scale=.5]{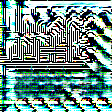}
	\end{subfigure}
	\begin{subfigure}{0.12\textwidth}
		\includegraphics[scale=.5]{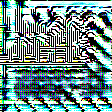}
	\end{subfigure}
	\begin{subfigure}{0.12\textwidth}
		\includegraphics[scale=.5]{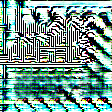}
	\end{subfigure} 
	\caption{C-DUP on UCF-101
	}
	\label{fig:v_CUP}
\end{figure*}

\begin{figure*}[t]
	\centering
	\begin{subfigure}{0.12\textwidth}
		\includegraphics[scale=.5]{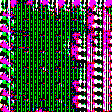}
	\end{subfigure}
	\begin{subfigure}{0.12\textwidth}
		\includegraphics[scale=.5]{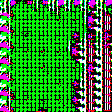}
	\end{subfigure}
	\begin{subfigure}{0.12\textwidth}
		\includegraphics[scale=.5]{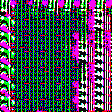}
	\end{subfigure} \vspace{0.01\textwidth}
	\begin{subfigure}{0.12\textwidth}
		\includegraphics[scale=.5]{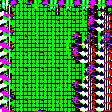}
	\end{subfigure}
	\begin{subfigure}{0.12\textwidth}
		\includegraphics[scale=.5]{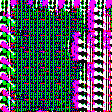}
	\end{subfigure}
	\begin{subfigure}{0.12\textwidth}
		\includegraphics[scale=.5]{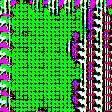}
	\end{subfigure}
	\begin{subfigure}{0.12\textwidth}
		\includegraphics[scale=.5]{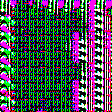}
	\end{subfigure}
	\begin{subfigure}{0.12\textwidth}
		\includegraphics[scale=.5]{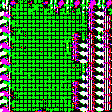} 
	\end{subfigure}
	\caption{C-DUP on Jester for $T_1 = \{\text{slding hand right}\}$}
	\label{fig:sliding_visual}
\end{figure*}

\begin{figure*}[t]
	\centering
	\begin{subfigure}{0.12\textwidth}
		\includegraphics[scale=.5]{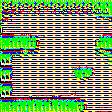}
	\end{subfigure}
	\begin{subfigure}{0.12\textwidth}
		\includegraphics[scale=.5]{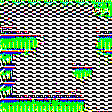}
	\end{subfigure}
	\begin{subfigure}{0.12\textwidth}
		\includegraphics[scale=.5]{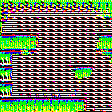}
	\end{subfigure} \vspace{0.01\textwidth}
	\begin{subfigure}{0.12\textwidth}
		\includegraphics[scale=.5]{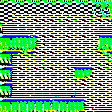}
	\end{subfigure}
	\begin{subfigure}{0.12\textwidth}
		\includegraphics[scale=.5]{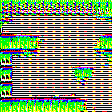}
	\end{subfigure}
	\begin{subfigure}{0.12\textwidth}
		\includegraphics[scale=.5]{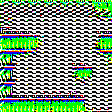}
	\end{subfigure}
	\begin{subfigure}{0.12\textwidth}
		\includegraphics[scale=.5]{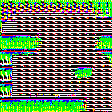}
	\end{subfigure}
	\begin{subfigure}{0.12\textwidth}
		\includegraphics[scale=.5]{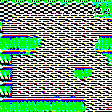} 
	\end{subfigure}
	\caption{C-DUP on Jester for $T_2 = \{\text{shaking hand}\}$}
	\label{fig:shake_visual}
\end{figure*}

\subsection{Experimental Setup}
\label{sec:exptdesc}
\heading{Discriminator set-up for our experiments:}
We used the C3D classifier as our discriminator.
To set up the discriminator for our experiments,
we pre-train the C3D classifier as described in \S \ref{sec:realtime}.
The discriminator is then used to train our generator.
For our experiments on the UCF101 dataset,
we use the C3D model available in the Github repository \cite{c3dgit}.
This pre-trained C3D model achieves an average clip classification
accuracy of $96.1\%$ on the UCF101 dataset in benign settings
(i.e., no adversarial inputs).

For the experiments on the Jester dataset,
we fine-tune the C3D model from the Github repository \cite{c3dgit}.
First, we change the output size of the last fully connected layer to 27,
since there are 27 gesture classes in Jester.
We use a learning rate with exponential decay \cite{zeiler2012adadelta} to train the model.
The starting learning rate for the last fully connected layer is set to be $10^{-3}$
and $10^{-4}$ for all the other layers.
The decay step is set to 600 and the decay rate is 0.9.
The fine-tuning phase is completed in 3 epochs and
we achieve a clip classification accuracy of $90.03\%$ in benign settings.

\heading{Generator set-up for our experiments:}
For building our generators, we refer to the generative model used by
Vondrik \etal \cite{vondrick2016generating},
which has 3D de-convolution layers.

For generators for both C-DUP and 2D-DUP,
we use five 3D de-convolution layers \cite{biggs20103d}.
The first four layers are followed by a batch normalization \cite{ioffe2015batch}
and a $ReLU$ \cite{nair2010rectified} activation function.
The last layer is followed by a $tanh$ \cite{kalman1992tanh} layer.
The kernel size for all 3D de-convolutions is set to be $3\times 3\times 3$.
To generate 3D perturbations (i.e., sequence of perturbation frames),
we set the kernel stride in the C-DUP generator to 1 in both
the spatial and temporal dimensions for the first layer,
and 2 in both the spatial and temporal dimensions for the following 4 layers.
To generate a single-frame 2D perturbation,
the kernel stride in the temporal dimension is set to 1
(i.e., 2D deconvolution) for all layers in the 2D-CUP generator,
and the spatial dimension stride is 1 for the first layer and 2 for the following layers.
The numbers of filters are shown in brackets in
\autoref{fig:arc_roll} and \autoref{fig:arc_tile}.
The input noise vector for both generators are sampled from a uniform distribution $U[-1,1]$
and the dimension of the noise vector is set to be 100.
For training both generators, we use a learning rate with exponential decay.
The starting learning rate is 0.002.
The decay step is 2000 and the decay rate is 0.95.
Unless otherwise specified, the weight balancing the two objectives,
i.e., $\lambda$, is set to 1 to reflect equal importance between
misclassifying the target class and retaining the correct
classification for all the other (non-target) classes.

\heading{Technical Implementation}:
All the models are implemented in TensorFlow~\cite{abadi2016tensorflow}
with the Adam optimizer \cite{kingma2014adam}.
Training was performed on 16 Tesla K80 GPU cards with the batch size set to 32.

\heading {Dataset setup for our experiments:}
On the UCF-101 dataset (denoted UCF-101 for short),
different sets of target class $T$ are tested.
We use $T=\{\text{apply lipstick}\}$ for presenting the results in the paper.
Experiments using other target sets also yield similar results.
UCF-101 has 101 classes of human actions in total.
The target set $T$ contains only one class while
the ``non-target'' set $S=X-T$ contains 100 classes.
The number of training inputs from the non-target classes is
approximately 100 times the number of training inputs from the target class.
Directly training with UCF-101 may cause a problem due to the imbalance
in the datasets containing the target and non-target classes \cite{longadge2013class}.
Therefore, we under-sample the non-target classes by a factor of 10.
Further, when loading a batch of inputs for training,
we fetch half the batch of inputs from the target set and
the other half from the non-target set in order to balance the inputs.

For the Jester dataset, we also choose different sets of target classes.
We use two target sets $T_1=\{\text{sliding hand right}\}$ and
$T_2=\{\text{shaking hands} \}$ as our representative examples
because they are exemplars of two different scenarios.
Since we seek to showcase an attack on a video classification system,
we care about how the perturbations affect both the appearance information
and temporal flow information, especially the latter.
For instance, the `sliding hand right' class has
a temporally similar class `sliding two fingers right;'
as a consequence, it may be easier for attackers to cause clips in the former class
to be misclassified as the later class
(because the temporal information does not need to be perturbed much).
On the other hand, `shaking hands' is not temporally similar to any other class.
Comparing the results of these two target sets could provide some
empirical evidence on the impact of the temporal flow on our perturbations.
Similar to UCF-101, the number of inputs from the non-target classes is
around 26 times the number of inputs from the target class
(since there are 27 classes in total and we only have one target class
in each experiment).
So we under-sample the non-target inputs by a factor of 4.
We also set up the environment to load half of the inputs from the
target set and the other half from the non-target set, in every batch.

\begin{figure*}[t]
	\centering
	\begin{subfigure}{0.4\textwidth}
		\includegraphics[scale=.4]{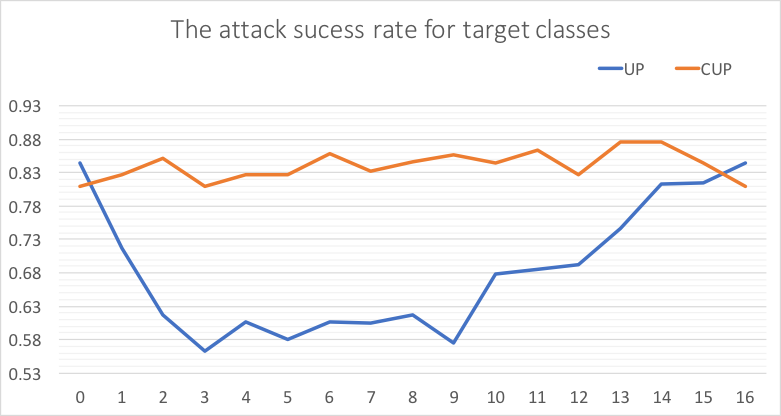}
		\caption{Attack success rate on UCF-101 for target class \\ 'applying lipstick'}
		\label{fig:ucf_target}
	\end{subfigure}
	\hspace{1mm}
	\begin{subfigure}{0.4\textwidth}
		\includegraphics[scale=.4]{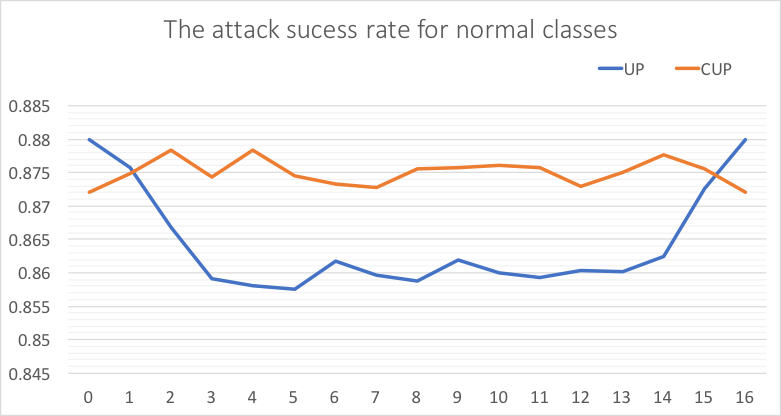}
		\caption{Attack success rate on UCF-101 for other non-target classes (all except 'applying lipstick')}
		\label{fig:ucf_non}
	\end{subfigure}
\hspace{1mm}
	\begin{subfigure}{0.4\textwidth}
        \includegraphics[scale=.2]{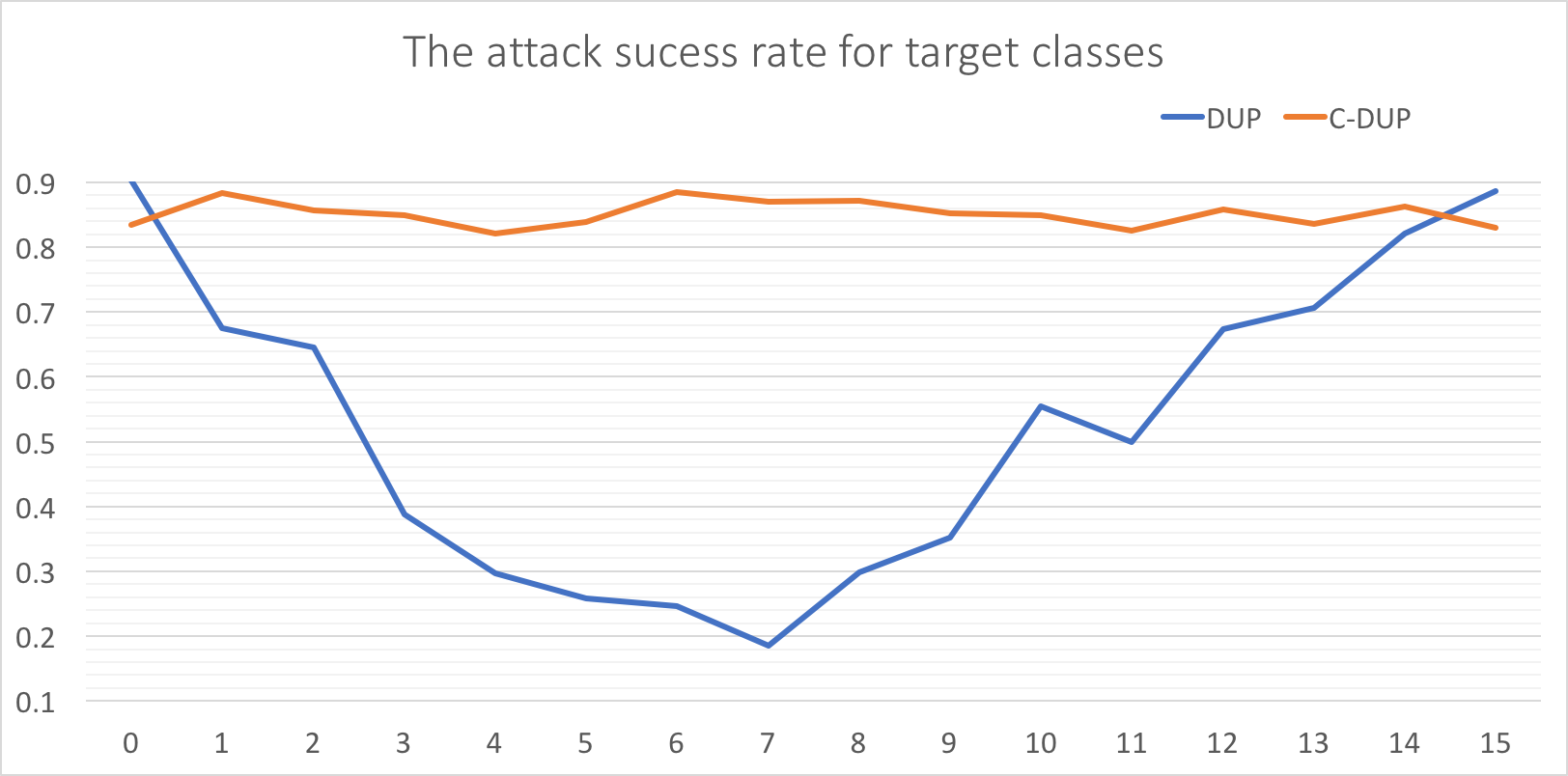}
        \caption{Attack success rate on Jester for target class \\ 'sliding hands right'}
        \label{fig:sliding_target}
    \end{subfigure}
\hspace{1mm}
    \begin{subfigure}{0.4\textwidth}
        \includegraphics[scale=.2]{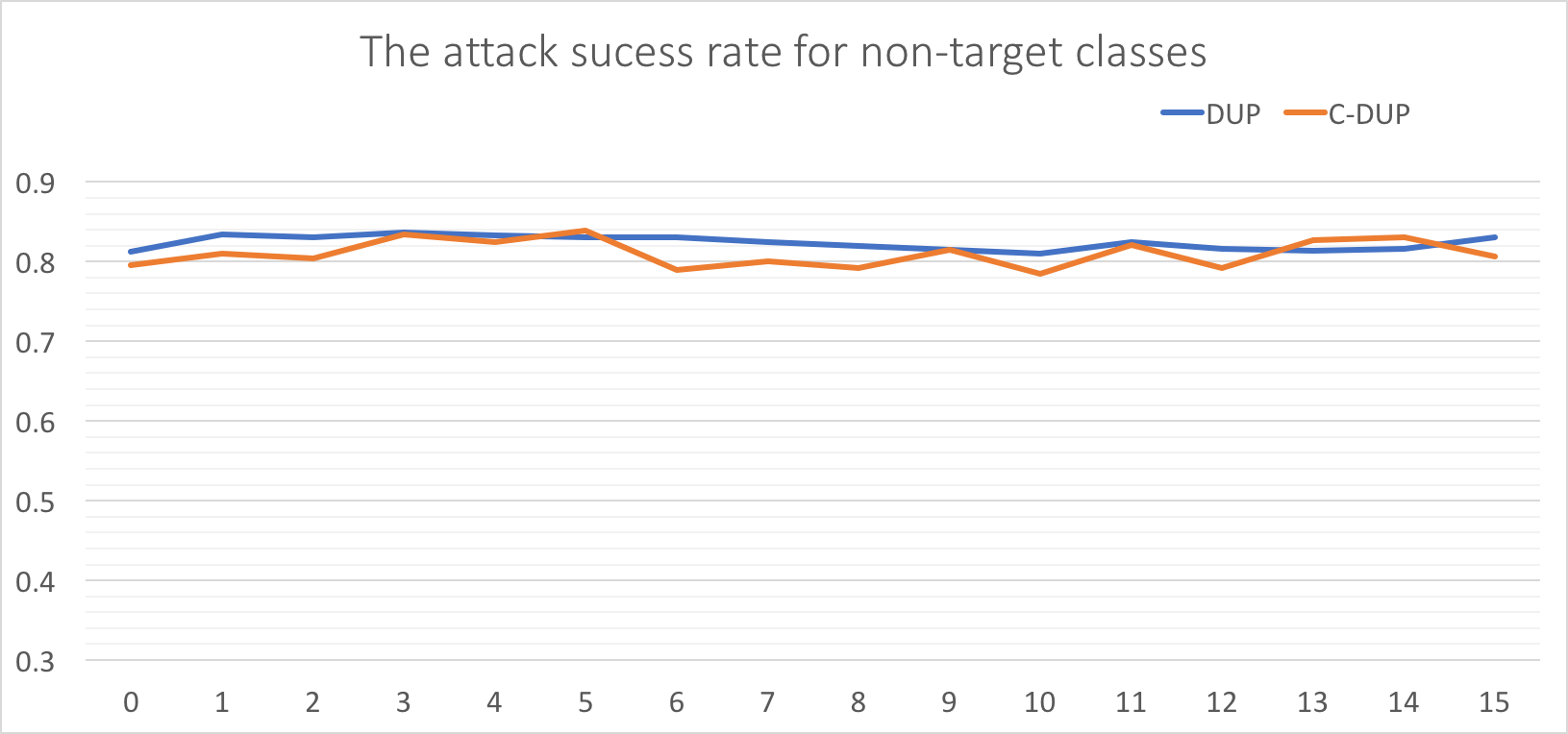}
        \caption{Attack success rate on Jester for non-target classes (all excepet 'sliding right')}
        \label{fig:sliding_non}
    \end{subfigure}
\hspace{1mm}
    \begin{subfigure}{0.4\textwidth}
        \includegraphics[scale=.2]{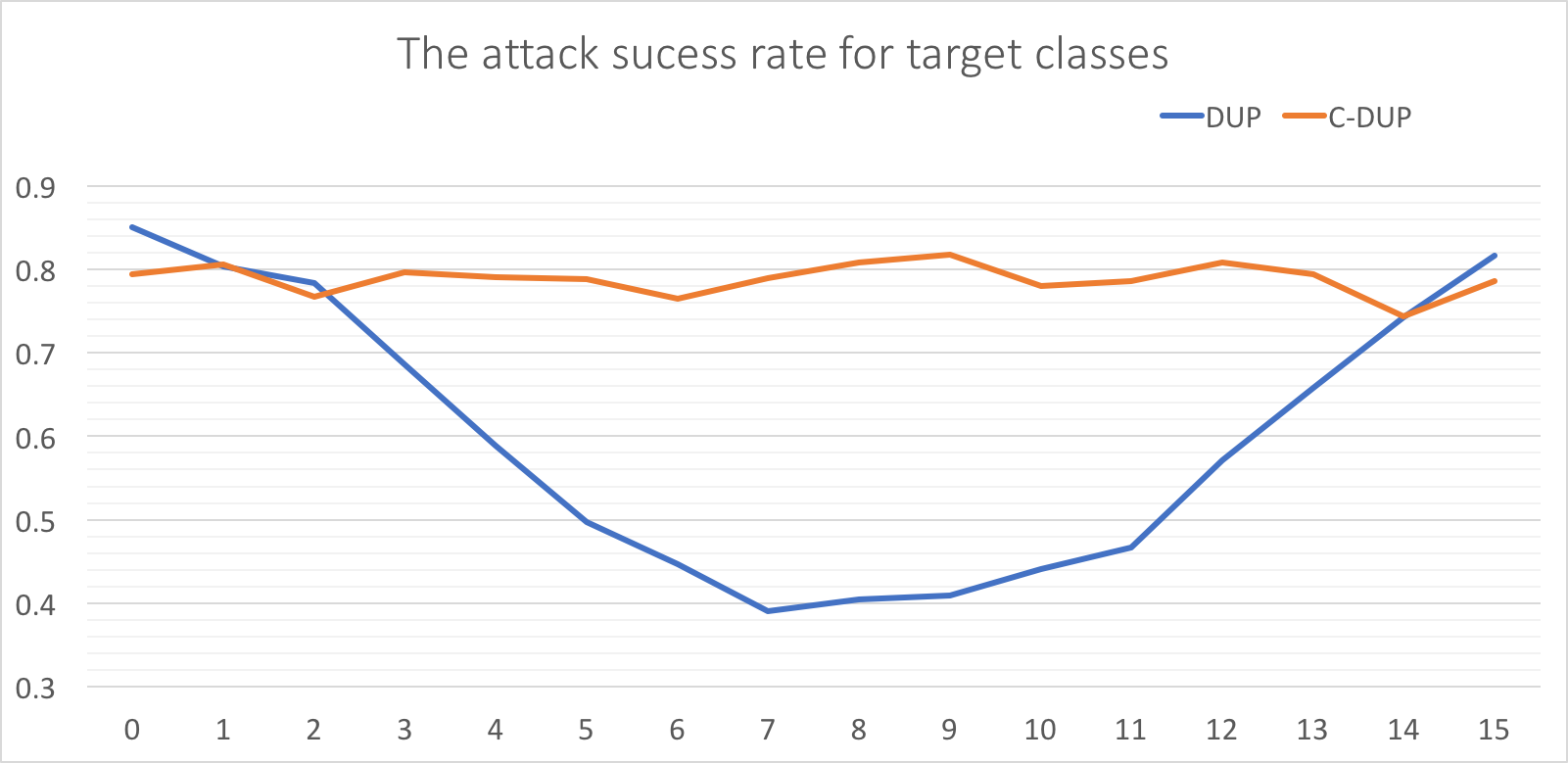}
        \caption{Attack success rate on Jester for target class \\ 'shaking hand'}
        \label{fig:shake_target}
    \end{subfigure}
\hspace{1mm}
    \begin{subfigure}{0.4\textwidth}
        \includegraphics[scale=.2]{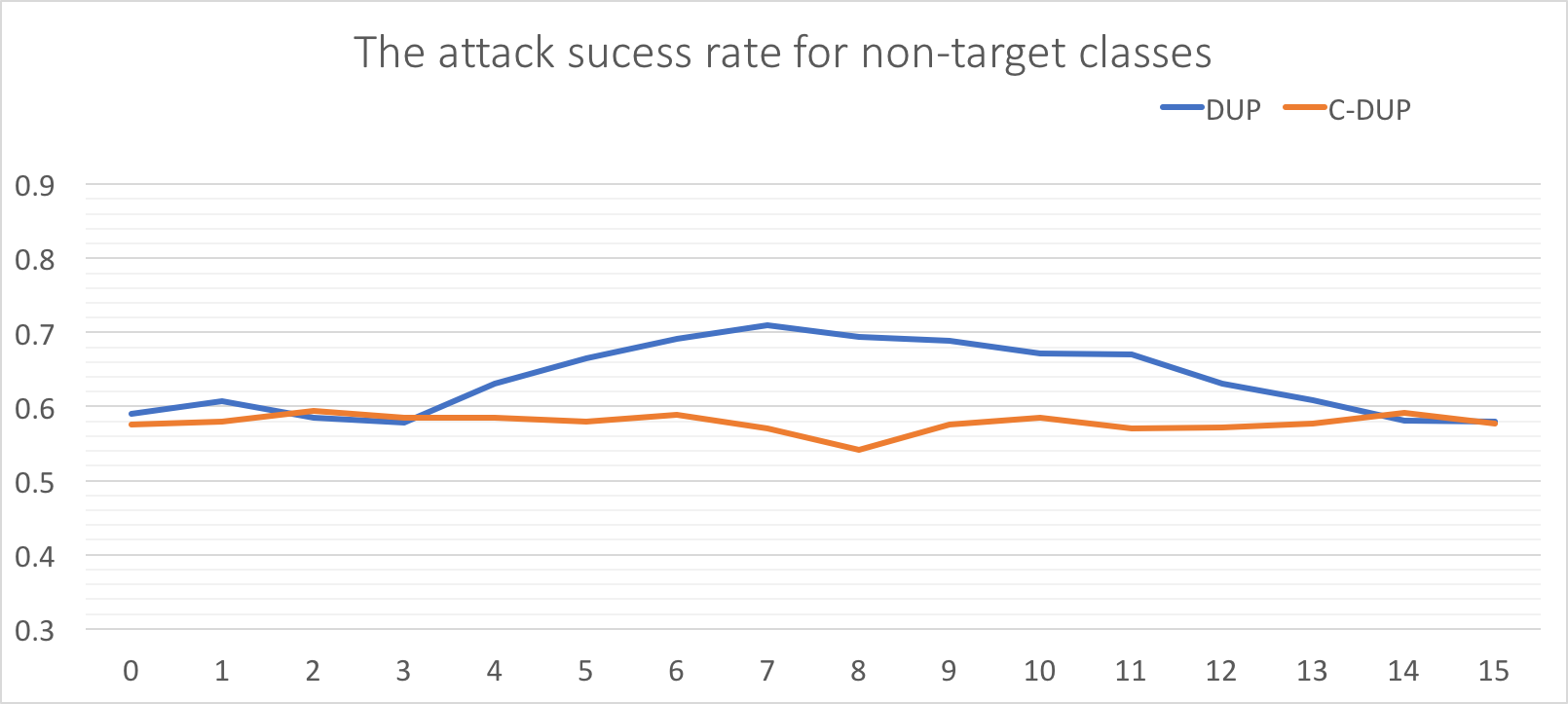}
        \caption{Attack success rate on Jester for non-target classes (all except 'shaking hand')}
        \label{fig:shake_non}
    \end{subfigure}
    \caption{attack success rate for DUP and C-DUP}
\end{figure*}

\vspace{-0.07in}
\heading{Metrics of interest:}
For measuring the efficacy of our perturbations, we consider two metrics.
\textit{First,} the perturbations added on the videos should be quasi-imperceptible.
\textit{Second}, the attack success rate for the target and the non-target classes should be high.
We define attack success rates as follows:

\squishlist
	\item The attack success rate for the target class is the misclassification rate;
	\item the attack success rate for the other classes is the correct classification rate.
\squishend

\subsection{C-DUP Perturbations}
\label{eval:cdup}

In this subsection, we discuss the results of the C-DUP perturbation attack.
We use the universal perturbations (UP) and DUP perturbations as our baselines.

\vspace{1mm}
\subsubsection{Experimental Results on UCF101}~
\vspace{-2mm}

\heading{Visualizing the perturbation:}
The perturbation clip generated by the DUP model is shown in \autoref{fig:v_DUP} and
the perturbation clip generated by C-DUP model is shown in \autoref{fig:v_CUP}.
The perturbation clip has 16 frames, and we
represent a visual representation of the first 8 frames for illustration\footnote{The complete 16-frame perturbation clips are shown in an appendix.}
We observe that the perturbation from DUP manifests an obvious disturbance among the frames.
With C-DUP, the perturbation frames look similar,
which implies that C-DUP does not perturb the temporal information by much, in UCF101.

\heading{Attack success rates:}
Recalling the discussion in \S \ref{sec:perturbframe}, one can expect that UP would cause inputs 
from the target class to be misclassified, but also significantly affect the correct classification 
of the other non-target inputs. On the other hand, oue would expect that DUP would achieve a stealthy attack, which would not cause much effect on the classification of non-target classes.

Based on the discussion in  \S \ref{sec:temporal},
we expect that DUP would work well when the perturbation clip is
well-aligned with the start point of each input clip to the classifier;
and the attack success rate would degrade as the misalignment increases.
We expect C-DUP would overcome this boundary effect and provide a better
overall attack performance (even with temporal misalignment).

We observe that when one considers no misalignment, DUP achieves a target misclassification rate
of 84.49 \%, while the non-target classes are correctly classified with a rate of 88.03 \%.
In contrast, when UP achieves a misclassification rate of 84.01 \% for the target class, only
45.2 \% of the other classes are correctly classified.
Given UP's inferior performance, we do not consider it any further in our evaluations.

The attack success rates with DUP and C-DUP, on the UCF-101 test set,
are shown in \autoref{fig:ucf_target} and \autoref{fig:ucf_non}.
The $x$ axis is the misalignment between perturbation clip and
and the input clip to the classifier. 
\autoref{fig:ucf_target} depicts the average misclassification rate for inputs from the target class.
We observe that when there is no misalignment, the attack success rate with the DUP is $84.49\%$,
which is in fact slightly higher than C-DUP.
However, the attack success rate with C-DUP is significantly higher
when there is misalignment.
Furthermore, the average attack success rate across all alignments for the target class with C-DUP is $84\%$,
while with DUP it is only $68.26\%$.
This demonstrates that C-DUP is more robust against boundary effects.

\autoref{fig:ucf_non} shows that, with regards to the classification of inputs from
the non-target classes, C-DUP also achieves a performance slightly better than DUP
when there is mismatch.
The average attack success rate (across all alignments) with C-DUP is $87.52\%$ here,
while with DUP it is $84.19\%$.

\begin{figure*}[t]
    \centering
    \begin{subfigure}{0.19\textwidth}
        \includegraphics[scale=.7]{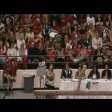}
    \end{subfigure}
    \begin{subfigure}{0.19\textwidth}
        \includegraphics[scale=.7]{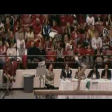}
    \end{subfigure}
    \begin{subfigure}{0.19\textwidth}
        \includegraphics[scale=.7]{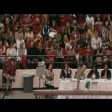}
    \end{subfigure}
    \begin{subfigure}{0.19\textwidth}
        \includegraphics[scale=.7]{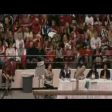}
    \end{subfigure}\vspace{0.02\textwidth}
    \begin{subfigure}{0.19\textwidth}
        \includegraphics[scale=.7]{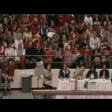}
    \end{subfigure}
    \begin{subfigure}{0.19\textwidth}
        \includegraphics[scale=.7]{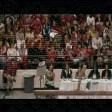}
    \end{subfigure}
    \begin{subfigure}{0.19\textwidth}
        \includegraphics[scale=.7]{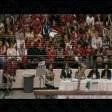}
    \end{subfigure}
    \begin{subfigure}{0.19\textwidth}
        \includegraphics[scale=.7]{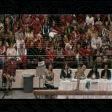}
    \end{subfigure}
    \begin{subfigure}{0.19\textwidth}
        \includegraphics[scale=.7]{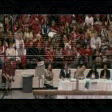}
    \end{subfigure}
    \begin{subfigure}{0.19\textwidth}
        \includegraphics[scale=.7]{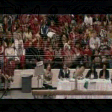}
    \end{subfigure}
    \caption{Visualizing images after adding 2D dual purpose universal perturbation: Original frames are displayed in the first row and perturbed frames are displayed in the second row. The perturbation added to the frames in the second row is mostly imperceptible to the human eye.}
    \label{fig:2D_CUP}
\end{figure*}

\vspace{1mm}
\subsubsection{Experimental Results on Jester}~
\vspace{-2mm}

\heading{Visualizing the perturbation:}
Visual representations of the C-DUP perturbations for the two target sets,
$T_1=\{\text{sliding hand right}\}$ and $T_2=\{\text{shaking hands} \}$
are shown in  \autoref{fig:sliding_visual} and \autoref{fig:shake_visual}.
We notice that compared to the perturbation
generated on UCF-101 (see \autoref{fig:v_CUP}).
there is a more pronounced evolution with respect to Jester
(see \autoref{fig:sliding_visual} and \autoref{fig:shake_visual}).
We conjecture that this is because UCF-101 is a coarse-grained action dataset
in which the spatial (appearance) information is dominant.
As a consequence, the C3D model does not extract/need much temporal
information to perform well. 
However, Jester is a fine-grained action dataset
where temporal information plays a more important role.
Therefore, in line with expectations, we find that in order to attack
the C3D model trained on the Jester dataset,
a higher extent of perturbations are required on the temporal dimension.


\heading{Attack success rate:}
To showcase a comparison of the misclassification rates with respect to
the target class between the two schemes (DUP and C-DUP),
we adjust the weighting factor $\lambda$ such that the classification
accuracy with respect to non-target classes are similar.
By choosing $\lambda$ = 1.5 for DUP and 1 for C-DUP,
we are able to achieve this.
The attack success rates for the above two target sets are shown in
\autoref{fig:sliding_target} and \autoref{fig:sliding_non},
{\em and} \autoref{fig:shake_target} and \autoref{fig:shake_non}, respectively.
We see that with respect to $T_1=\{\text{sliding hand right}\}$,
the results are similar to what we observe with UCF101.
The attack success rates for C-DUP are a little lower than DUP when
the offset is 0. This is to be expected since DUP is tailored for this specific offset.
However, C-DUP outperforms DUP when there is a misalignment.
The average success rate for C-DUP is $85.14\%$ for the target class
and $81.03\%$ for the other (non-target) classes.
The average success rate for DUP is $52.42\%$ for the target class
and $82.36\%$ for the other (non-target) classes.

Next we consider the case with $T_2=\{\text{shaking hands} \}$.
In general, we find that both DUP and C-DUP achieve relatively
lower success rates especially with regards to the other (non-target) classes.
As discussed in  \S \ref{sec:exptdesc}, unlike in the previous case
where `sliding two fingers right' is temporally similar to `sliding hand right',
no other class is temporally similar to `shaking hand'.
Therefore it is harder to achieve misclassification.
\autoref{fig:shake_non} depicts the attack success rates for non-target classes.
The average success rate for non-target classes are similar
(because of the bias in $\lambda$ as discussed earlier)
and in fact slightly higher for DUP (a finer bias could make them exactly the same).
The attack success rates with the two approaches for the target class 
are shown in \autoref{fig:shake_target}.
We see that C-DUP significantly outperforms DUP in terms of misclassification
efficacy because of its robustness to temporal misalignment.
The average attack success rate for the target class with C-DUP is $79.03\%$
while for DUP it is only $57.78\%$.
Overall, our C-DUP outperforms DUP in being able to achieve a better
misclassification rate for the target class.
We believe that although stealth is affected to some extent,
it is still reasonably high.

\subsection{2D-Dual Purpose Universal Perturbations}

The visual representations of the perturbations with C-DUP show that
perturbations on all the frames are visually similar.
Thus, we ask if it is possible to add ``the same perturbation'' on every frame
and still achieve a successful attack.
In other words, will the 2D-DUP perturbation attack yield performance
similar to the C-DUP attack.

{

\vspace{1mm}
\subsubsection{Experimental Results on the UCF101 Dataset}~\\
\vspace{-4mm}

\heading{Visual impact of the perturbation:}
We present a sequence of original frames and its corresponding perturbed
frames in \autoref{fig:2D_CUP}.
Original frames are displayed in the first row and perturbed frames are
displayed in the second row.
We observe that the perturbation added on the frames is
quasi-imperceptible to human eyes (similar results are seen with C-DUP {\color{black} and
are presented in an appendix}).

\heading{Attack success rate:}
By adding 2D-DUP on the video clip,
we achieve an attack success rate of $87.58\%$ with respect to the target class
and an attack success rate of $83.37\%$ for the non-target classes.
Recall that the average attack success rates with C-DUP were $87.52\%$ and $84.00\%$, respectively.
Thus, the performance of 2D-DUP seems to be on par with that of C-DUP on the UCF101 dataset.
This demonstrates that C3D is vulnerable even if the same 2D
perturbation generated by our approach is added on to every frame.

\vspace{1mm}
\subsubsection{Experimental Results on Jester Dataset}~\\
\vspace{-4mm}

\heading{Attack success rate:}
For $T_1=\{\text{sliding hand right}\}$,
the attack success rate for the target class is $84.64\%$ and the
attack success rate for the non-target classes is $80.04\%$.
This shows that 2D-DUP is also successful on some target classes in the fine-grained,
Jester action dataset.

For the target set $T_2$, the success rate for the target class drops to $70.92\%$,
while the success rate for non-target class is $54.83\%$.
This is slightly degraded compared to the success rates achieved with C-DUP
($79.03\%$ and $57.78\%$ respectively),
but is still reasonable.
This degradation is due to more significant temporal changes in this case
(unlike in the case of $T_1$) and a single 2D perturbation is
less effective in manipulating these changes.
In contrast, because the perturbations evolve with C-DUP,
they are much more effective in achieving the misclassification of the target class.

\section{Discussion}

\heading{Black box attacks:}
In this work we assumed that the adversary is fully aware of the DNN being deployed
(i.e., white box attacks).
However, in practice the adversary may need to determine the type of
DNN being used in the video classification system,
and so a black box approach may be needed.
Given recent studies on the transferability of adversarial inputs~\cite{papernot2016transferability},
we believe black box attacks are also feasible.
We will explore this in our future work.

\heading{Context dependency:}
Second, the approach that we developed does not account for contextual information,
i.e., consistency between the misclassified result and the context.
For example, the context relates to a baseball game a human overseeing
the system may notice an inconsistency
when the action of hitting a ball is misclassified into applying makeup.
Similarly, because of context, if there is a series of actions that we want to misclassify,
inconsistency in the misclassification results
(e.g., different actions across the clips) may also raise an alarm.
For example, let us consider a case where the actions include running,
kicking a ball, and applying make up.
While the first two actions can be considered to be {\em reasonable}
with regards to appearing together in a video, the latter two are unlikely.
Generating perturbations that are consistent with the context of the video
is a line of future work that we will explore and is likely to require
new techniques.
In fact, looking for consistency in context may be a potential defense,
and we will also examine this in depth in the future.

\heading{Defenses:}
{\color{black} In order to defense against the attacks against video classification systems,
one can try some existing defense methods in image area, such as feature squeezing \cite{xu2017feature1,xu2017feature2} and ensemble adversarial training \cite{tramer2017ensemble} (although their
effectiveness is yet unknown). Considering the properties of video that were discussed, 
we envision some exclusive defense methods for protecting video classification systems
below, which we will explore in future work.}

One approach is to examine the consistency between the classification of
consecutive frames (considered as images) within a clip,
and between consecutive clips in a stream. 
A sudden change in the classification results could raise an alarm.
However, while this defense will work well in cases where the temporal flow
is not pronounced (e.g., the UCF101 dataset),
it may not work well in cases with pronounced temporal flows.
For example, with respect to the Jester dataset,
with just an image it may be hard to determine whether the hand is being moved right or left.

The second line of defense may be to identify an object that is present in the video,
e.g., a soccer ball in a video clip that depicts a kicking action.
We can use an additional classifier to identify such objects
in the individual frames that compose the video.
Then, we can look for consistency with regards to the action and the object,
e.g., a kicking action is can be associated with a soccer ball,
but cannot be associated with a make up kit.
Towards realizing this line of defense,
we could use existing image classifiers in conjunction with
the video classification system.
We will explore this in future work.

{

\section{Related Work}

There is quite a bit of work \cite{biggio2013evasion,biggio2014pattern,huang2011adversarial} on
investigating the vulnerability of machine learning systems to adversarial inputs. 
Researchers have shown that generally, small magnitude perturbations added to input samples, 
change the predictions made by machine learning models. 
Most efforts, however, do not consider real-time temporally varying inputs such as video.
Unlike these efforts, our study is focused on the generation of adversarial perturbations to fool DNN based real-time video action recognition systems.

The threat of adversarial samples to deep-learning systems has also received considerable attention recently. There are several papers in the literature (e.g., \cite{moosavi2017universal,moosavi2016deepfool,sharif2016accessorize, goodfellow2014generative,goodfellow2014explaining}) that have shown that 
the state-of-the-art DNN based learning systems are also vulnerable to well-designed adversarial 
perturbations \cite{szegedy2013intriguing}. Szegedy \etal show that the addition of 
hardly perceptible perturbation on an image, can cause a neural network to misclassify the image. 
Goodfellow \etal \cite{goodfellow2014explaining} analyze the potency of adversarial samples available in the physical world, in terms of fooling neural networks. 
Moosavi-Dezfooli \etal \cite{moosavi2017universal, moosavi2016deepfool,mopuri2017fast} make a significant
contribution by generating image-agnostic perturbations, which they call universal adversarial perturbations. 
These perturbations can cause all natural images belonging to target classes to be misclassified 
with high probability.  


GANs or generative adversarial networks have been employed by Goodfellow \etal \cite{goodfellow2014generative} and Radford \etal \cite{radford2015unsupervised} in generating natural images. 
Mopuri \etal \cite{mopuri2017nag} extend a GAN architecture to train a generator to model 
universal perturbations for images. Their objective was to explore the space of
the distribution of universal adversarial perturbations in the image space. 
We significantly extend the generative framework introduced by Mopuri \etal \cite{mopuri2017nag}. 
In addition, unlike their work which focused on generating adversarial perturbations for images, our study focuses on the generation of effective perturbations to attack videos.

The feasibility of adversarial attacks against other types
of learning systems including face-recognition systems \cite{sharif2016accessorize,mccoyd2016spoofing,sharif2017adversarial}, 
voice recognition systems \cite{carlini2016hidden} and malware classification systems \cite{grosse2016adversarial} has been studied. However, these studies do not account for the unique input characteristics 
that are present in real-time video activity recognition systems.

\section{Conclusions}

In this paper, we investigate the problem of generating adversarial samples for attacking video classification systems. We identify three key challenges that will need to be addressed in order to generate such samples namely, generating perturbations in real-time, making the perturbations stealthy and accounting for the temporal structure of frames in a video clip. We exploit recent advances in GAN architectures, extending them significantly to solve these challenges and generate very potent adversarial samples against video classification systems. We perform extensive experiments on two different datasets one of which captures coarse-grained actions (e.g., applying make up) while the other captures fine-grained actions (hand gestures). We demonstrate that our approaches are extremely potent, achieving around 80 \% attack success rates in both cases. We also discuss possible defenses that we propose to investigate in future work.


\appendix
\section{Appendix: Visualizing perturbations on UCF101} 

\begin{figure*}[!h]
	\centering
	\begin{subfigure}{0.12\textwidth}
		\includegraphics[scale=.5]{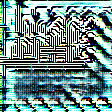}
	\end{subfigure}
	\begin{subfigure}{0.12\textwidth}
		\includegraphics[scale=.5]{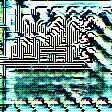}
	\end{subfigure}
	\begin{subfigure}{0.12\textwidth}
		\includegraphics[scale=.5]{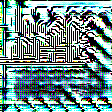}
	\end{subfigure} \vspace{0.01\textwidth}
	\begin{subfigure}{0.12\textwidth}
		\includegraphics[scale=.5]{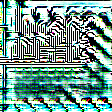}
	\end{subfigure}
	\begin{subfigure}{0.12\textwidth}
		\includegraphics[scale=.5]{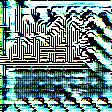}
	\end{subfigure}
	\begin{subfigure}{0.12\textwidth}
		\includegraphics[scale=.5]{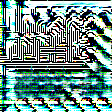}
	\end{subfigure}
	\begin{subfigure}{0.12\textwidth}
		\includegraphics[scale=.5]{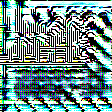}
	\end{subfigure}
	\begin{subfigure}{0.12\textwidth}
		\includegraphics[scale=.5]{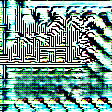} 
	\end{subfigure}
	\begin{subfigure}{0.12\textwidth}
		\includegraphics[scale=.5]{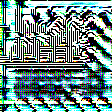}
	\end{subfigure}
	\begin{subfigure}{0.12\textwidth}
		\includegraphics[scale=.5]{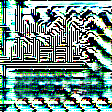} 
	\end{subfigure}
	\begin{subfigure}{0.12\textwidth}
		\includegraphics[scale=.5]{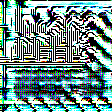}
	\end{subfigure}
	\begin{subfigure}{0.12\textwidth}
		\includegraphics[scale=.5]{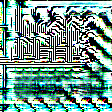}
	\end{subfigure}
	\begin{subfigure}{0.12\textwidth}
		\includegraphics[scale=.5]{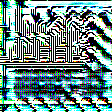}
	\end{subfigure}
	\begin{subfigure}{0.12\textwidth}
		\includegraphics[scale=.5]{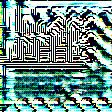}
	\end{subfigure}
	\begin{subfigure}{0.12\textwidth}
		\includegraphics[scale=.5]{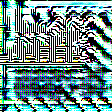}
	\end{subfigure}
	\begin{subfigure}{0.12\textwidth}
		\includegraphics[scale=.5]{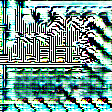}
	\end{subfigure}
	\caption{Circular Dual-Purpose Universal Perturbation (C-DUP) for UCF101}
	\label{fig:a_ucf_c_dup}
\end{figure*}

We visualize a full range of perturbations and their corresponding video representations 
on the UCF-101 dataset here.
\autoref{fig:a_ucf_c_dup} shows the entire Circular Dual-Purpose Universal Perturbation (C-DUP) clip 
(with 16 frames) generated for a target class $T=\{\text{apply lipstick}\}$. 
Similarly,
\autoref{fig:a_ucf_2d} shows the entire 2D Dual-Purpose Universal Perturbation (2D-DUP) for 
same target class.

\autoref{fig:a_ucf_org} shows a clean video clip from UCF101 dataset.
\autoref{fig:a_ucf_C_p} shows the same video clip perturbed by the C-DUP shown in \autoref{fig:a_ucf_c_dup}. 
Similarly,
\autoref{fig:a_ucf_C_2d} shows the video clip perturbed by 2D-DUP shown in \autoref{fig:a_ucf_2d}. 
In both cases, we observe that the perturbation is quasi-imperceptible to human vision i.e., the
frames in the videos look very much like the original.

\begin{figure*}[!h]
	\centering
	\begin{subfigure}{0.12\textwidth}
		\includegraphics[scale=.5]{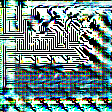}
	\end{subfigure}
	\begin{subfigure}{0.12\textwidth}
		\includegraphics[scale=.5]{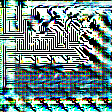}
	\end{subfigure}
	\begin{subfigure}{0.12\textwidth}
		\includegraphics[scale=.5]{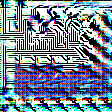}
	\end{subfigure} \vspace{0.01\textwidth}
	\begin{subfigure}{0.12\textwidth}
		\includegraphics[scale=.5]{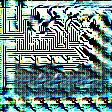}
	\end{subfigure}
	\begin{subfigure}{0.12\textwidth}
		\includegraphics[scale=.5]{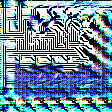}
	\end{subfigure}
	\begin{subfigure}{0.12\textwidth}
		\includegraphics[scale=.5]{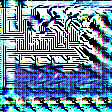}
	\end{subfigure}
	\begin{subfigure}{0.12\textwidth}
		\includegraphics[scale=.5]{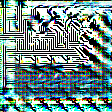}
	\end{subfigure}
	\begin{subfigure}{0.12\textwidth}
		\includegraphics[scale=.5]{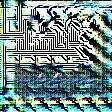} 
	\end{subfigure}
	\begin{subfigure}{0.12\textwidth}
		\includegraphics[scale=.5]{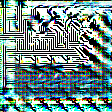}
	\end{subfigure}
	\begin{subfigure}{0.12\textwidth}
		\includegraphics[scale=.5]{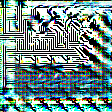} 
	\end{subfigure}
	\begin{subfigure}{0.12\textwidth}
		\includegraphics[scale=.5]{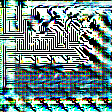}
	\end{subfigure}
	\begin{subfigure}{0.12\textwidth}
		\includegraphics[scale=.5]{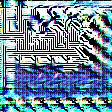}
	\end{subfigure}
	\begin{subfigure}{0.12\textwidth}
		\includegraphics[scale=.5]{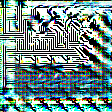}
	\end{subfigure}
	\begin{subfigure}{0.12\textwidth}
		\includegraphics[scale=.5]{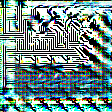}
	\end{subfigure}
	\begin{subfigure}{0.12\textwidth}
		\includegraphics[scale=.5]{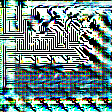}
	\end{subfigure}
	\begin{subfigure}{0.12\textwidth}
		\includegraphics[scale=.5]{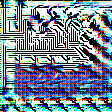}
	\end{subfigure}
	\caption{2D Dual-Purpose Universal Perturbation for UCF-101}
	\label{fig:a_ucf_2d}
\end{figure*}

\begin{figure*}[!h]
	\centering
	\begin{subfigure}{0.12\textwidth}
		\includegraphics[scale=.5]{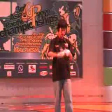}
	\end{subfigure}
	\begin{subfigure}{0.12\textwidth}
		\includegraphics[scale=.5]{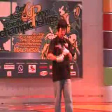}
	\end{subfigure}
	\begin{subfigure}{0.12\textwidth}
		\includegraphics[scale=.5]{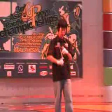}
	\end{subfigure} \vspace{0.01\textwidth}
	\begin{subfigure}{0.12\textwidth}
		\includegraphics[scale=.5]{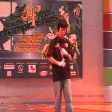}
	\end{subfigure}
	\begin{subfigure}{0.12\textwidth}
		\includegraphics[scale=.5]{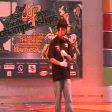}
	\end{subfigure}
	\begin{subfigure}{0.12\textwidth}
		\includegraphics[scale=.5]{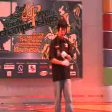}
	\end{subfigure}
	\begin{subfigure}{0.12\textwidth}
		\includegraphics[scale=.5]{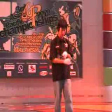}
	\end{subfigure}
	\begin{subfigure}{0.12\textwidth}
		\includegraphics[scale=.5]{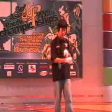} 
	\end{subfigure}
	\begin{subfigure}{0.12\textwidth}
		\includegraphics[scale=.5]{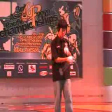}
	\end{subfigure}
	\begin{subfigure}{0.12\textwidth}
		\includegraphics[scale=.5]{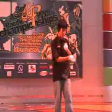} 
	\end{subfigure}
	\begin{subfigure}{0.12\textwidth}
		\includegraphics[scale=.5]{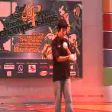}
	\end{subfigure}
	\begin{subfigure}{0.12\textwidth}
		\includegraphics[scale=.5]{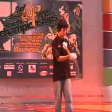}
	\end{subfigure}
	\begin{subfigure}{0.12\textwidth}
		\includegraphics[scale=.5]{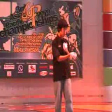}
	\end{subfigure}
	\begin{subfigure}{0.12\textwidth}
		\includegraphics[scale=.5]{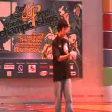}
	\end{subfigure}
	\begin{subfigure}{0.12\textwidth}
		\includegraphics[scale=.5]{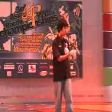}
	\end{subfigure}
	\begin{subfigure}{0.12\textwidth}
		\includegraphics[scale=.5]{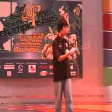}
	\end{subfigure}
	\caption{Original frames from UCF-101}
	\label{fig:a_ucf_org}
\end{figure*}

\begin{figure*}[!h]
	\centering
	\begin{subfigure}{0.12\textwidth}
		\includegraphics[scale=.5]{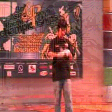}
	\end{subfigure}
	\begin{subfigure}{0.12\textwidth}
		\includegraphics[scale=.5]{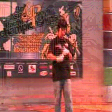}
	\end{subfigure}
	\begin{subfigure}{0.12\textwidth}
		\includegraphics[scale=.5]{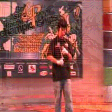}
	\end{subfigure} \vspace{0.01\textwidth}
	\begin{subfigure}{0.12\textwidth}
		\includegraphics[scale=.5]{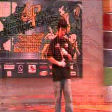}
	\end{subfigure}
	\begin{subfigure}{0.12\textwidth}
		\includegraphics[scale=.5]{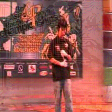}
	\end{subfigure}
	\begin{subfigure}{0.12\textwidth}
		\includegraphics[scale=.5]{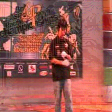}
	\end{subfigure}
	\begin{subfigure}{0.12\textwidth}
		\includegraphics[scale=.5]{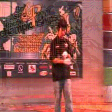}
	\end{subfigure}
	\begin{subfigure}{0.12\textwidth}
		\includegraphics[scale=.5]{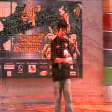} 
	\end{subfigure}
	\begin{subfigure}{0.12\textwidth}
		\includegraphics[scale=.5]{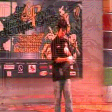}
	\end{subfigure}
	\begin{subfigure}{0.12\textwidth}
		\includegraphics[scale=.5]{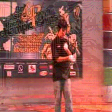} 
	\end{subfigure}
	\begin{subfigure}{0.12\textwidth}
		\includegraphics[scale=.5]{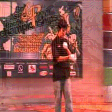}
	\end{subfigure}
	\begin{subfigure}{0.12\textwidth}
		\includegraphics[scale=.5]{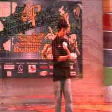}
	\end{subfigure}
	\begin{subfigure}{0.12\textwidth}
		\includegraphics[scale=.5]{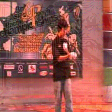}
	\end{subfigure}
	\begin{subfigure}{0.12\textwidth}
		\includegraphics[scale=.5]{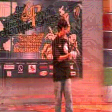}
	\end{subfigure}
	\begin{subfigure}{0.12\textwidth}
		\includegraphics[scale=.5]{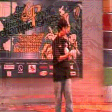}
	\end{subfigure}
	\begin{subfigure}{0.12\textwidth}
		\includegraphics[scale=.5]{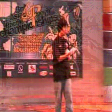}
	\end{subfigure}
	\caption{C-DUP perturbed frames from UCF-101}
	\label{fig:a_ucf_C_p}
\end{figure*}

\begin{figure*}[!h]
	\centering
	\begin{subfigure}{0.12\textwidth}
		\includegraphics[scale=.5]{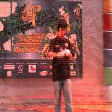}
	\end{subfigure}
	\begin{subfigure}{0.12\textwidth}
		\includegraphics[scale=.5]{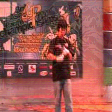}
	\end{subfigure}
	\begin{subfigure}{0.12\textwidth}
		\includegraphics[scale=.5]{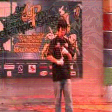}
	\end{subfigure} \vspace{0.01\textwidth}
	\begin{subfigure}{0.12\textwidth}
		\includegraphics[scale=.5]{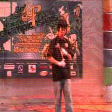}
	\end{subfigure}
	\begin{subfigure}{0.12\textwidth}
		\includegraphics[scale=.5]{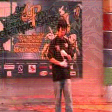}
	\end{subfigure}
	\begin{subfigure}{0.12\textwidth}
		\includegraphics[scale=.5]{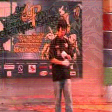}
	\end{subfigure}
	\begin{subfigure}{0.12\textwidth}
		\includegraphics[scale=.5]{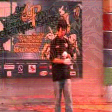}
	\end{subfigure}
	\begin{subfigure}{0.12\textwidth}
		\includegraphics[scale=.5]{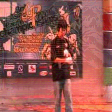} 
	\end{subfigure}
	\begin{subfigure}{0.12\textwidth}
		\includegraphics[scale=.5]{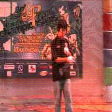}
	\end{subfigure}
	\begin{subfigure}{0.12\textwidth}
		\includegraphics[scale=.5]{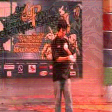} 
	\end{subfigure}
	\begin{subfigure}{0.12\textwidth}
		\includegraphics[scale=.5]{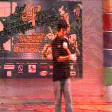}
	\end{subfigure}
	\begin{subfigure}{0.12\textwidth}
		\includegraphics[scale=.5]{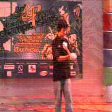}
	\end{subfigure}
	\begin{subfigure}{0.12\textwidth}
		\includegraphics[scale=.5]{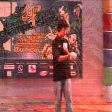}
	\end{subfigure}
	\begin{subfigure}{0.12\textwidth}
		\includegraphics[scale=.5]{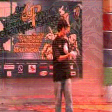}
	\end{subfigure}
	\begin{subfigure}{0.12\textwidth}
		\includegraphics[scale=.5]{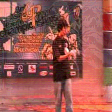}
	\end{subfigure}
	\begin{subfigure}{0.12\textwidth}
		\includegraphics[scale=.5]{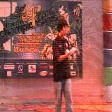}
	\end{subfigure}
	\caption{2D-DUP perturbed frames from UCF-101}
	\label{fig:a_ucf_C_2d}
\end{figure*}

\section{Appendix: Visualizing perturbations on Jester}

\autoref{fig:a_j_c} shows the entire C-DUP generated for clip to misclassify the target class $T=\{\text{shaking hand}\}$ on the Jester dataset. Similarly, \autoref{fig:a_j_2d}  depicts the entire 2D-DUP generated 
for misclassifying the same target class.

\autoref{fig:a_jester_o} shows a clean video clip from Jester dataset corresponding to 
shaking of the hand (visible only in some of the frames shown due to cropping).
\autoref{fig:a_jester_c} shows the clip perturbed by C-DUP shown in \autoref{fig:a_j_c} from 
the Jester dataset. 
Similarly \autoref{fig:a_jester_2d} shows the same clip perturbed by 2D-DUP shown in \autoref{fig:a_j_2d}. 
It is easy to see that the perturbation is quasi-imperceptible to human vision.

\begin{figure*}[!h]
	\centering
	\begin{subfigure}{0.12\textwidth}
		\includegraphics[scale=.5]{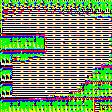}
	\end{subfigure}
	\begin{subfigure}{0.12\textwidth}
		\includegraphics[scale=.5]{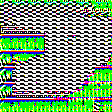}
	\end{subfigure}
	\begin{subfigure}{0.12\textwidth}
		\includegraphics[scale=.5]{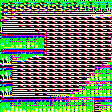}
	\end{subfigure} \vspace{0.01\textwidth}
	\begin{subfigure}{0.12\textwidth}
		\includegraphics[scale=.5]{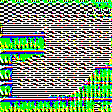}
	\end{subfigure}
	\begin{subfigure}{0.12\textwidth}
		\includegraphics[scale=.5]{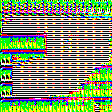}
	\end{subfigure}
	\begin{subfigure}{0.12\textwidth}
		\includegraphics[scale=.5]{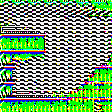}
	\end{subfigure}
	\begin{subfigure}{0.12\textwidth}
		\includegraphics[scale=.5]{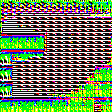}
	\end{subfigure}
	\begin{subfigure}{0.12\textwidth}
		\includegraphics[scale=.5]{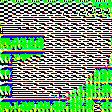} 
	\end{subfigure}
	\begin{subfigure}{0.12\textwidth}
		\includegraphics[scale=.5]{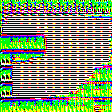}
	\end{subfigure}
	\begin{subfigure}{0.12\textwidth}
		\includegraphics[scale=.5]{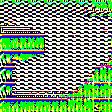} 
	\end{subfigure}
	\begin{subfigure}{0.12\textwidth}
		\includegraphics[scale=.5]{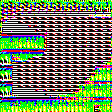}
	\end{subfigure}
	\begin{subfigure}{0.12\textwidth}
		\includegraphics[scale=.5]{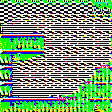}
	\end{subfigure}
	\begin{subfigure}{0.12\textwidth}
		\includegraphics[scale=.5]{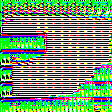}
	\end{subfigure}
	\begin{subfigure}{0.12\textwidth}
		\includegraphics[scale=.5]{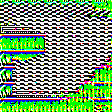}
	\end{subfigure}
	\begin{subfigure}{0.12\textwidth}
		\includegraphics[scale=.5]{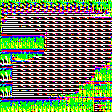}
	\end{subfigure}
	\begin{subfigure}{0.12\textwidth}
		\includegraphics[scale=.5]{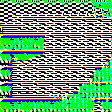}
	\end{subfigure}
	\caption{C-DUP for Jester}
	\label{fig:a_j_c}
\end{figure*}

\begin{figure*}[!h]
	\centering
	\begin{subfigure}{0.12\textwidth}
		\includegraphics[scale=.5]{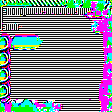}
	\end{subfigure}
	\begin{subfigure}{0.12\textwidth}
		\includegraphics[scale=.5]{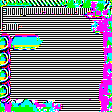}
	\end{subfigure}
	\begin{subfigure}{0.12\textwidth}
		\includegraphics[scale=.5]{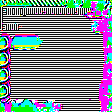}
	\end{subfigure} \vspace{0.01\textwidth}
	\begin{subfigure}{0.12\textwidth}
		\includegraphics[scale=.5]{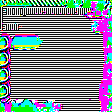}
	\end{subfigure}
	\begin{subfigure}{0.12\textwidth}
		\includegraphics[scale=.5]{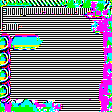}
	\end{subfigure}
	\begin{subfigure}{0.12\textwidth}
		\includegraphics[scale=.5]{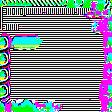}
	\end{subfigure}
	\begin{subfigure}{0.12\textwidth}
		\includegraphics[scale=.5]{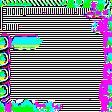}
	\end{subfigure}
	\begin{subfigure}{0.12\textwidth}
		\includegraphics[scale=.5]{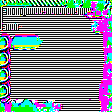} 
	\end{subfigure}
	\begin{subfigure}{0.12\textwidth}
		\includegraphics[scale=.5]{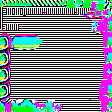}
	\end{subfigure}
	\begin{subfigure}{0.12\textwidth}
		\includegraphics[scale=.5]{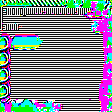} 
	\end{subfigure}
	\begin{subfigure}{0.12\textwidth}
		\includegraphics[scale=.5]{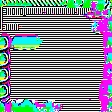}
	\end{subfigure}
	\begin{subfigure}{0.12\textwidth}
		\includegraphics[scale=.5]{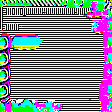}
	\end{subfigure}
	\begin{subfigure}{0.12\textwidth}
		\includegraphics[scale=.5]{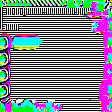}
	\end{subfigure}
	\begin{subfigure}{0.12\textwidth}
		\includegraphics[scale=.5]{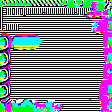}
	\end{subfigure}
	\begin{subfigure}{0.12\textwidth}
		\includegraphics[scale=.5]{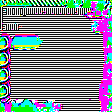}
	\end{subfigure}
	\begin{subfigure}{0.12\textwidth}
		\includegraphics[scale=.5]{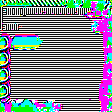}
	\end{subfigure}
	\caption{2D-DUP for Jester}
	\label{fig:a_j_2d}
\end{figure*}

\begin{figure*}[!h]
	\centering
	\begin{subfigure}{0.12\textwidth}
		\includegraphics[scale=.5]{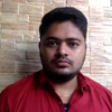}
	\end{subfigure}
	\begin{subfigure}{0.12\textwidth}
		\includegraphics[scale=.5]{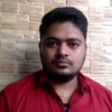}
	\end{subfigure}
	\begin{subfigure}{0.12\textwidth}
		\includegraphics[scale=.5]{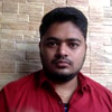}
	\end{subfigure} \vspace{0.01\textwidth}
	\begin{subfigure}{0.12\textwidth}
		\includegraphics[scale=.5]{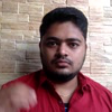}
	\end{subfigure}
	\begin{subfigure}{0.12\textwidth}
		\includegraphics[scale=.5]{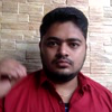}
	\end{subfigure}
	\begin{subfigure}{0.12\textwidth}
		\includegraphics[scale=.5]{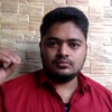}
	\end{subfigure}
	\begin{subfigure}{0.12\textwidth}
		\includegraphics[scale=.5]{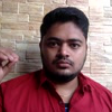}
	\end{subfigure}
	\begin{subfigure}{0.12\textwidth}
		\includegraphics[scale=.5]{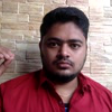} 
	\end{subfigure}
	\begin{subfigure}{0.12\textwidth}
		\includegraphics[scale=.5]{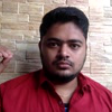}
	\end{subfigure}
	\begin{subfigure}{0.12\textwidth}
		\includegraphics[scale=.5]{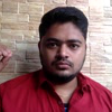} 
	\end{subfigure}
	\begin{subfigure}{0.12\textwidth}
		\includegraphics[scale=.5]{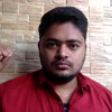}
	\end{subfigure}
	\begin{subfigure}{0.12\textwidth}
		\includegraphics[scale=.5]{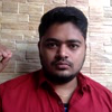}
	\end{subfigure}
	\begin{subfigure}{0.12\textwidth}
		\includegraphics[scale=.5]{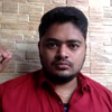}
	\end{subfigure}
	\begin{subfigure}{0.12\textwidth}
		\includegraphics[scale=.5]{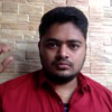}
	\end{subfigure}
	\begin{subfigure}{0.12\textwidth}
		\includegraphics[scale=.5]{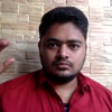}
	\end{subfigure}
	\begin{subfigure}{0.12\textwidth}
		\includegraphics[scale=.5]{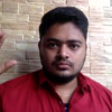}
	\end{subfigure}
	\caption{Original frames from Jester}
	\label{fig:a_jester_o}
\end{figure*}

\begin{figure*}[!h]
	\centering
	\begin{subfigure}{0.12\textwidth}
		\includegraphics[scale=.5]{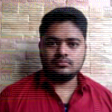}
	\end{subfigure}
	\begin{subfigure}{0.12\textwidth}
		\includegraphics[scale=.5]{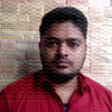}
	\end{subfigure}
	\begin{subfigure}{0.12\textwidth}
		\includegraphics[scale=.5]{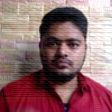}
	\end{subfigure} \vspace{0.01\textwidth}
	\begin{subfigure}{0.12\textwidth}
		\includegraphics[scale=.5]{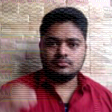}
	\end{subfigure}
	\begin{subfigure}{0.12\textwidth}
		\includegraphics[scale=.5]{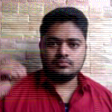}
	\end{subfigure}
	\begin{subfigure}{0.12\textwidth}
		\includegraphics[scale=.5]{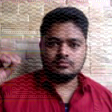}
	\end{subfigure}
	\begin{subfigure}{0.12\textwidth}
		\includegraphics[scale=.5]{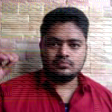}
	\end{subfigure}
	\begin{subfigure}{0.12\textwidth}
		\includegraphics[scale=.5]{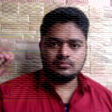} 
	\end{subfigure}
	\begin{subfigure}{0.12\textwidth}
		\includegraphics[scale=.5]{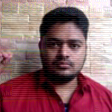}
	\end{subfigure}
	\begin{subfigure}{0.12\textwidth}
		\includegraphics[scale=.5]{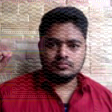} 
	\end{subfigure}
	\begin{subfigure}{0.12\textwidth}
		\includegraphics[scale=.5]{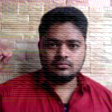}
	\end{subfigure}
	\begin{subfigure}{0.12\textwidth}
		\includegraphics[scale=.5]{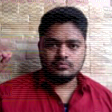}
	\end{subfigure}
	\begin{subfigure}{0.12\textwidth}
		\includegraphics[scale=.5]{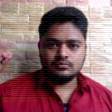}
	\end{subfigure}
	\begin{subfigure}{0.12\textwidth}
		\includegraphics[scale=.5]{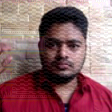}
	\end{subfigure}
	\begin{subfigure}{0.12\textwidth}
		\includegraphics[scale=.5]{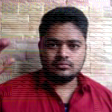}
	\end{subfigure}
	\begin{subfigure}{0.12\textwidth}
		\includegraphics[scale=.5]{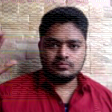}
	\end{subfigure}
	\caption{C-DUP perturbed frames from Jester}
	\label{fig:a_jester_c}
\end{figure*}

\begin{figure*}[!h]
	\centering
	\begin{subfigure}{0.12\textwidth}
		\includegraphics[scale=.5]{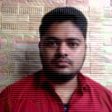}
	\end{subfigure}
	\begin{subfigure}{0.12\textwidth}
		\includegraphics[scale=.5]{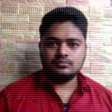}
	\end{subfigure}
	\begin{subfigure}{0.12\textwidth}
		\includegraphics[scale=.5]{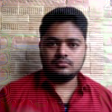}
	\end{subfigure} \vspace{0.01\textwidth}
	\begin{subfigure}{0.12\textwidth}
		\includegraphics[scale=.5]{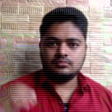}
	\end{subfigure}
	\begin{subfigure}{0.12\textwidth}
		\includegraphics[scale=.5]{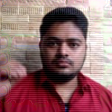}
	\end{subfigure}
	\begin{subfigure}{0.12\textwidth}
		\includegraphics[scale=.5]{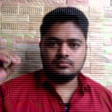}
	\end{subfigure}
	\begin{subfigure}{0.12\textwidth}
		\includegraphics[scale=.5]{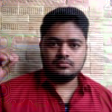}
	\end{subfigure}
	\begin{subfigure}{0.12\textwidth}
		\includegraphics[scale=.5]{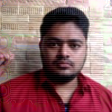} 
	\end{subfigure}
	\begin{subfigure}{0.12\textwidth}
		\includegraphics[scale=.5]{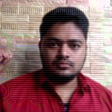}
	\end{subfigure}
	\begin{subfigure}{0.12\textwidth}
		\includegraphics[scale=.5]{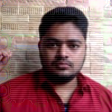} 
	\end{subfigure}
	\begin{subfigure}{0.12\textwidth}
		\includegraphics[scale=.5]{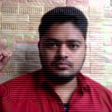}
	\end{subfigure}
	\begin{subfigure}{0.12\textwidth}
		\includegraphics[scale=.5]{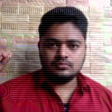}
	\end{subfigure}
	\begin{subfigure}{0.12\textwidth}
		\includegraphics[scale=.5]{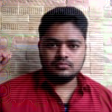}
	\end{subfigure}
	\begin{subfigure}{0.12\textwidth}
		\includegraphics[scale=.5]{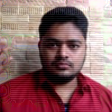}
	\end{subfigure}
	\begin{subfigure}{0.12\textwidth}
		\includegraphics[scale=.5]{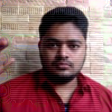}
	\end{subfigure}
	\begin{subfigure}{0.12\textwidth}
		\includegraphics[scale=.5]{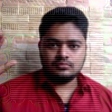}
	\end{subfigure}
	\caption{2D-DUP perturbed frames from Jester}
	\label{fig:a_jester_2d}
\end{figure*}

\bibliographystyle{ACM-Reference-Format}
\bibliography{references}


\begin{thebibliography}{53}


\ifx \showCODEN    \undefined \def \showCODEN     #1{\unskip}     \fi
\ifx \showDOI      \undefined \def \showDOI       #1{#1}\fi
\ifx \showISBNx    \undefined \def \showISBNx     #1{\unskip}     \fi
\ifx \showISBNxiii \undefined \def \showISBNxiii  #1{\unskip}     \fi
\ifx \showISSN     \undefined \def \showISSN      #1{\unskip}     \fi
\ifx \showLCCN     \undefined \def \showLCCN      #1{\unskip}     \fi
\ifx \shownote     \undefined \def \shownote      #1{#1}          \fi
\ifx \showarticletitle \undefined \def \showarticletitle #1{#1}   \fi
\ifx \showURL      \undefined \def \showURL       {\relax}        \fi
\providecommand\bibfield[2]{#2}
\providecommand\bibinfo[2]{#2}
\providecommand\natexlab[1]{#1}
\providecommand\showeprint[2][]{arXiv:#2}

\bibitem[\protect\citeauthoryear{Abadi, Barham, Chen, Chen, Davis, Dean, Devin,
  Ghemawat, Irving, Isard, et~al\mbox{.}}{Abadi et~al\mbox{.}}{2016}]%
        {abadi2016tensorflow}
\bibfield{author}{\bibinfo{person}{Mart{\'\i}n Abadi}, \bibinfo{person}{Paul
  Barham}, \bibinfo{person}{Jianmin Chen}, \bibinfo{person}{Zhifeng Chen},
  \bibinfo{person}{Andy Davis}, \bibinfo{person}{Jeffrey Dean},
  \bibinfo{person}{Matthieu Devin}, \bibinfo{person}{Sanjay Ghemawat},
  \bibinfo{person}{Geoffrey Irving}, \bibinfo{person}{Michael Isard},
  {et~al\mbox{.}}} \bibinfo{year}{2016}\natexlab{}.
\newblock \showarticletitle{TensorFlow: A System for Large-Scale Machine
  Learning.}. In \bibinfo{booktitle}{\emph{OSDI}}, Vol.~\bibinfo{volume}{16}.
  \bibinfo{pages}{265--283}.
\newblock


\bibitem[\protect\citeauthoryear{Biggio, Corona, Maiorca, Nelson,
  {\v{S}}rndi{\'c}, Laskov, Giacinto, and Roli}{Biggio et~al\mbox{.}}{2013}]%
        {biggio2013evasion}
\bibfield{author}{\bibinfo{person}{Battista Biggio}, \bibinfo{person}{Igino
  Corona}, \bibinfo{person}{Davide Maiorca}, \bibinfo{person}{Blaine Nelson},
  \bibinfo{person}{Nedim {\v{S}}rndi{\'c}}, \bibinfo{person}{Pavel Laskov},
  \bibinfo{person}{Giorgio Giacinto}, {and} \bibinfo{person}{Fabio Roli}.}
  \bibinfo{year}{2013}\natexlab{}.
\newblock \showarticletitle{Evasion attacks against machine learning at test
  time}. In \bibinfo{booktitle}{\emph{Joint European conference on machine
  learning and knowledge discovery in databases}}. Springer,
  \bibinfo{pages}{387--402}.
\newblock


\bibitem[\protect\citeauthoryear{Biggio, Fumera, and Roli}{Biggio
  et~al\mbox{.}}{2014}]%
        {biggio2014pattern}
\bibfield{author}{\bibinfo{person}{Battista Biggio}, \bibinfo{person}{Giorgio
  Fumera}, {and} \bibinfo{person}{Fabio Roli}.}
  \bibinfo{year}{2014}\natexlab{}.
\newblock \showarticletitle{Pattern recognition systems under attack: Design
  issues and research challenges}.
\newblock \bibinfo{journal}{\emph{International Journal of Pattern Recognition
  and Artificial Intelligence}} \bibinfo{volume}{28}, \bibinfo{number}{07}
  (\bibinfo{year}{2014}), \bibinfo{pages}{1460002}.
\newblock


\bibitem[\protect\citeauthoryear{Biggs}{Biggs}{2010}]%
        {biggs20103d}
\bibfield{author}{\bibinfo{person}{David~SC Biggs}.}
  \bibinfo{year}{2010}\natexlab{}.
\newblock \showarticletitle{3D deconvolution microscopy}.
\newblock \bibinfo{journal}{\emph{Current Protocols in Cytometry}}
  (\bibinfo{year}{2010}), \bibinfo{pages}{12--19}.
\newblock


\bibitem[\protect\citeauthoryear{Carlini, Mishra, Vaidya, Zhang, Sherr,
  Shields, Wagner, and Zhou}{Carlini et~al\mbox{.}}{2016}]%
        {carlini2016hidden}
\bibfield{author}{\bibinfo{person}{Nicholas Carlini}, \bibinfo{person}{Pratyush
  Mishra}, \bibinfo{person}{Tavish Vaidya}, \bibinfo{person}{Yuankai Zhang},
  \bibinfo{person}{Micah Sherr}, \bibinfo{person}{Clay Shields},
  \bibinfo{person}{David Wagner}, {and} \bibinfo{person}{Wenchao Zhou}.}
  \bibinfo{year}{2016}\natexlab{}.
\newblock \showarticletitle{Hidden Voice Commands.}. In
  \bibinfo{booktitle}{\emph{USENIX Security Symposium}}.
  \bibinfo{pages}{513--530}.
\newblock


\bibitem[\protect\citeauthoryear{Carreira and Zisserman}{Carreira and
  Zisserman}{2017}]%
        {carreira2017quo}
\bibfield{author}{\bibinfo{person}{Joao Carreira} {and} \bibinfo{person}{Andrew
  Zisserman}.} \bibinfo{year}{2017}\natexlab{}.
\newblock \showarticletitle{Quo vadis, action recognition? a new model and the
  kinetics dataset}. In \bibinfo{booktitle}{\emph{2017 IEEE Conference on
  Computer Vision and Pattern Recognition (CVPR)}}. IEEE,
  \bibinfo{pages}{4724--4733}.
\newblock


\bibitem[\protect\citeauthoryear{Dataset}{Dataset}{2016}]%
        {jester2016}
\bibfield{author}{\bibinfo{person}{Jester Dataset}.}
  \bibinfo{year}{2016}\natexlab{}.
\newblock \bibinfo{title}{{Humans performing pre-defined hand actions}}.
\newblock \bibinfo{howpublished}{\url{https://20bn.com/datasets/jester}}.
\newblock
\newblock
\shownote{[Online; accessed 30-April-2018].}


\bibitem[\protect\citeauthoryear{Fanello, Gori, Metta, and Odone}{Fanello
  et~al\mbox{.}}{2013}]%
        {fanello2013one}
\bibfield{author}{\bibinfo{person}{Sean~Ryan Fanello}, \bibinfo{person}{Ilaria
  Gori}, \bibinfo{person}{Giorgio Metta}, {and} \bibinfo{person}{Francesca
  Odone}.} \bibinfo{year}{2013}\natexlab{}.
\newblock \showarticletitle{One-shot learning for real-time action
  recognition}. In \bibinfo{booktitle}{\emph{Iberian Conference on Pattern
  Recognition and Image Analysis}}. Springer, \bibinfo{pages}{31--40}.
\newblock


\bibitem[\protect\citeauthoryear{Foroughi, Aski, and Pourreza}{Foroughi
  et~al\mbox{.}}{2008}]%
        {foroughi2008intelligent}
\bibfield{author}{\bibinfo{person}{Homa Foroughi},
  \bibinfo{person}{Baharak~Shakeri Aski}, {and} \bibinfo{person}{Hamidreza
  Pourreza}.} \bibinfo{year}{2008}\natexlab{}.
\newblock \showarticletitle{Intelligent video surveillance for monitoring fall
  detection of elderly in home environments}. In
  \bibinfo{booktitle}{\emph{Computer and Information Technology, 2008. ICCIT
  2008. 11th International Conference on}}. IEEE, \bibinfo{pages}{219--224}.
\newblock


\bibitem[\protect\citeauthoryear{Goodfellow, Pouget-Abadie, Mirza, Xu,
  Warde-Farley, Ozair, Courville, and Bengio}{Goodfellow
  et~al\mbox{.}}{2014a}]%
        {goodfellow2014generative}
\bibfield{author}{\bibinfo{person}{Ian Goodfellow}, \bibinfo{person}{Jean
  Pouget-Abadie}, \bibinfo{person}{Mehdi Mirza}, \bibinfo{person}{Bing Xu},
  \bibinfo{person}{David Warde-Farley}, \bibinfo{person}{Sherjil Ozair},
  \bibinfo{person}{Aaron Courville}, {and} \bibinfo{person}{Yoshua Bengio}.}
  \bibinfo{year}{2014}\natexlab{a}.
\newblock \showarticletitle{Generative adversarial nets}. In
  \bibinfo{booktitle}{\emph{Advances in neural information processing
  systems}}. \bibinfo{pages}{2672--2680}.
\newblock


\bibitem[\protect\citeauthoryear{Goodfellow, Shlens, and Szegedy}{Goodfellow
  et~al\mbox{.}}{2014b}]%
        {goodfellow2014explaining}
\bibfield{author}{\bibinfo{person}{Ian~J Goodfellow}, \bibinfo{person}{Jonathon
  Shlens}, {and} \bibinfo{person}{Christian Szegedy}.}
  \bibinfo{year}{2014}\natexlab{b}.
\newblock \showarticletitle{Explaining and harnessing adversarial examples}.
\newblock \bibinfo{journal}{\emph{arXiv preprint arXiv:1412.6572}}
  (\bibinfo{year}{2014}).
\newblock


\bibitem[\protect\citeauthoryear{Grosse, Papernot, Manoharan, Backes, and
  McDaniel}{Grosse et~al\mbox{.}}{2016}]%
        {grosse2016adversarial}
\bibfield{author}{\bibinfo{person}{Kathrin Grosse}, \bibinfo{person}{Nicolas
  Papernot}, \bibinfo{person}{Praveen Manoharan}, \bibinfo{person}{Michael
  Backes}, {and} \bibinfo{person}{Patrick McDaniel}.}
  \bibinfo{year}{2016}\natexlab{}.
\newblock \showarticletitle{Adversarial perturbations against deep neural
  networks for malware classification}.
\newblock \bibinfo{journal}{\emph{arXiv preprint arXiv:1606.04435}}
  (\bibinfo{year}{2016}).
\newblock


\bibitem[\protect\citeauthoryear{Gu, Wang, Kuen, Ma, Shahroudy, Shuai, Liu,
  Wang, Wang, Cai, et~al\mbox{.}}{Gu et~al\mbox{.}}{2017}]%
        {gu2017recent}
\bibfield{author}{\bibinfo{person}{Jiuxiang Gu}, \bibinfo{person}{Zhenhua
  Wang}, \bibinfo{person}{Jason Kuen}, \bibinfo{person}{Lianyang Ma},
  \bibinfo{person}{Amir Shahroudy}, \bibinfo{person}{Bing Shuai},
  \bibinfo{person}{Ting Liu}, \bibinfo{person}{Xingxing Wang},
  \bibinfo{person}{Gang Wang}, \bibinfo{person}{Jianfei Cai}, {et~al\mbox{.}}}
  \bibinfo{year}{2017}\natexlab{}.
\newblock \showarticletitle{Recent advances in convolutional neural networks}.
\newblock \bibinfo{journal}{\emph{Pattern Recognition}} (\bibinfo{year}{2017}).
\newblock


\bibitem[\protect\citeauthoryear{Herath, Harandi, and Porikli}{Herath
  et~al\mbox{.}}{2017}]%
        {herath2017going}
\bibfield{author}{\bibinfo{person}{Samitha Herath}, \bibinfo{person}{Mehrtash
  Harandi}, {and} \bibinfo{person}{Fatih Porikli}.}
  \bibinfo{year}{2017}\natexlab{}.
\newblock \showarticletitle{Going deeper into action recognition: A survey}.
\newblock \bibinfo{journal}{\emph{Image and vision computing}}
  \bibinfo{volume}{60} (\bibinfo{year}{2017}), \bibinfo{pages}{4--21}.
\newblock


\bibitem[\protect\citeauthoryear{Huang, Joseph, Nelson, Rubinstein, and
  Tygar}{Huang et~al\mbox{.}}{2011}]%
        {huang2011adversarial}
\bibfield{author}{\bibinfo{person}{Ling Huang}, \bibinfo{person}{Anthony~D
  Joseph}, \bibinfo{person}{Blaine Nelson}, \bibinfo{person}{Benjamin~IP
  Rubinstein}, {and} \bibinfo{person}{JD Tygar}.}
  \bibinfo{year}{2011}\natexlab{}.
\newblock \showarticletitle{Adversarial machine learning}. In
  \bibinfo{booktitle}{\emph{Proceedings of the 4th ACM workshop on Security and
  artificial intelligence}}. ACM, \bibinfo{pages}{43--58}.
\newblock


\bibitem[\protect\citeauthoryear{Huang, Li, Poursaeed, Hopcroft, and
  Belongie}{Huang et~al\mbox{.}}{2017}]%
        {huang2017stacked}
\bibfield{author}{\bibinfo{person}{Xun Huang}, \bibinfo{person}{Yixuan Li},
  \bibinfo{person}{Omid Poursaeed}, \bibinfo{person}{John Hopcroft}, {and}
  \bibinfo{person}{Serge Belongie}.} \bibinfo{year}{2017}\natexlab{}.
\newblock \showarticletitle{Stacked generative adversarial networks}. In
  \bibinfo{booktitle}{\emph{IEEE Conference on Computer Vision and Pattern
  Recognition (CVPR)}}, Vol.~\bibinfo{volume}{2}. \bibinfo{pages}{4}.
\newblock


\bibitem[\protect\citeauthoryear{Ioffe and Szegedy}{Ioffe and Szegedy}{2015}]%
        {ioffe2015batch}
\bibfield{author}{\bibinfo{person}{Sergey Ioffe} {and}
  \bibinfo{person}{Christian Szegedy}.} \bibinfo{year}{2015}\natexlab{}.
\newblock \showarticletitle{Batch normalization: Accelerating deep network
  training by reducing internal covariate shift}.
\newblock \bibinfo{journal}{\emph{arXiv preprint arXiv:1502.03167}}
  (\bibinfo{year}{2015}).
\newblock


\bibitem[\protect\citeauthoryear{Kalman and Kwasny}{Kalman and Kwasny}{1992}]%
        {kalman1992tanh}
\bibfield{author}{\bibinfo{person}{Barry~L Kalman} {and}
  \bibinfo{person}{Stan~C Kwasny}.} \bibinfo{year}{1992}\natexlab{}.
\newblock \showarticletitle{Why tanh: choosing a sigmoidal function}. In
  \bibinfo{booktitle}{\emph{Neural Networks, 1992. IJCNN., International Joint
  Conference on}}, Vol.~\bibinfo{volume}{4}. IEEE, \bibinfo{pages}{578--581}.
\newblock


\bibitem[\protect\citeauthoryear{Karpathy, Toderici, Shetty, Leung, Sukthankar,
  and Fei-Fei}{Karpathy et~al\mbox{.}}{2014}]%
        {karpathy2014large}
\bibfield{author}{\bibinfo{person}{Andrej Karpathy}, \bibinfo{person}{George
  Toderici}, \bibinfo{person}{Sanketh Shetty}, \bibinfo{person}{Thomas Leung},
  \bibinfo{person}{Rahul Sukthankar}, {and} \bibinfo{person}{Li Fei-Fei}.}
  \bibinfo{year}{2014}\natexlab{}.
\newblock \showarticletitle{Large-scale video classification with convolutional
  neural networks}. In \bibinfo{booktitle}{\emph{Proceedings of the IEEE
  conference on Computer Vision and Pattern Recognition}}.
  \bibinfo{pages}{1725--1732}.
\newblock


\bibitem[\protect\citeauthoryear{Kataoka, Satoh, Aoki, Oikawa, and
  Matsui}{Kataoka et~al\mbox{.}}{2018a}]%
        {kataoka2018temporal}
\bibfield{author}{\bibinfo{person}{Hirokatsu Kataoka}, \bibinfo{person}{Yutaka
  Satoh}, \bibinfo{person}{Yoshimitsu Aoki}, \bibinfo{person}{Shoko Oikawa},
  {and} \bibinfo{person}{Yasuhiro Matsui}.} \bibinfo{year}{2018}\natexlab{a}.
\newblock \showarticletitle{Temporal and fine-grained pedestrian action
  recognition on driving recorder database}.
\newblock \bibinfo{journal}{\emph{Sensors}} \bibinfo{volume}{18},
  \bibinfo{number}{2} (\bibinfo{year}{2018}), \bibinfo{pages}{627}.
\newblock


\bibitem[\protect\citeauthoryear{Kataoka, Suzuki, Oikawa, Matsui, and
  Satoh}{Kataoka et~al\mbox{.}}{2018b}]%
        {kataoka2018drive}
\bibfield{author}{\bibinfo{person}{Hirokatsu Kataoka}, \bibinfo{person}{Teppei
  Suzuki}, \bibinfo{person}{Shoko Oikawa}, \bibinfo{person}{Yasuhiro Matsui},
  {and} \bibinfo{person}{Yutaka Satoh}.} \bibinfo{year}{2018}\natexlab{b}.
\newblock \showarticletitle{Drive Video Analysis for the Detection of Traffic
  Near-Miss Incidents}.
\newblock \bibinfo{journal}{\emph{arXiv preprint arXiv:1804.02555}}
  (\bibinfo{year}{2018}).
\newblock


\bibitem[\protect\citeauthoryear{Kingma and Ba}{Kingma and Ba}{2014}]%
        {kingma2014adam}
\bibfield{author}{\bibinfo{person}{Diederik~P Kingma} {and}
  \bibinfo{person}{Jimmy Ba}.} \bibinfo{year}{2014}\natexlab{}.
\newblock \showarticletitle{Adam: A method for stochastic optimization}.
\newblock \bibinfo{journal}{\emph{arXiv preprint arXiv:1412.6980}}
  (\bibinfo{year}{2014}).
\newblock


\bibitem[\protect\citeauthoryear{Lab}{Lab}{2014}]%
        {mitm2014videos}
\bibfield{author}{\bibinfo{person}{Kaspersky Lab}.}
  \bibinfo{year}{Defcon,2014}\natexlab{}.
\newblock \bibinfo{title}{{Man-in-the-middle attack on video surveillance
  systems}}.
\newblock
  \bibinfo{howpublished}{\url{https://securelist.com/does-cctv-put-the-public-at-risk-of-cyberattack/70008/}}.
\newblock
\newblock
\shownote{[Online; accessed 30-April-2018].}


\bibitem[\protect\citeauthoryear{Longadge and Dongre}{Longadge and
  Dongre}{2013}]%
        {longadge2013class}
\bibfield{author}{\bibinfo{person}{Rushi Longadge} {and}
  \bibinfo{person}{Snehalata Dongre}.} \bibinfo{year}{2013}\natexlab{}.
\newblock \showarticletitle{Class imbalance problem in data mining review}.
\newblock \bibinfo{journal}{\emph{arXiv preprint arXiv:1305.1707}}
  (\bibinfo{year}{2013}).
\newblock


\bibitem[\protect\citeauthoryear{McCoyd and Wagner}{McCoyd and Wagner}{2016}]%
        {mccoyd2016spoofing}
\bibfield{author}{\bibinfo{person}{Michael McCoyd} {and} \bibinfo{person}{David
  Wagner}.} \bibinfo{year}{2016}\natexlab{}.
\newblock \showarticletitle{Spoofing 2D Face Detection: Machines See People Who
  Aren't There}.
\newblock \bibinfo{journal}{\emph{arXiv preprint arXiv:1608.02128}}
  (\bibinfo{year}{2016}).
\newblock


\bibitem[\protect\citeauthoryear{Moosavi-Dezfooli, Fawzi, Fawzi, and
  Frossard}{Moosavi-Dezfooli et~al\mbox{.}}{2017}]%
        {moosavi2017universal}
\bibfield{author}{\bibinfo{person}{Seyed-Mohsen Moosavi-Dezfooli},
  \bibinfo{person}{Alhussein Fawzi}, \bibinfo{person}{Omar Fawzi}, {and}
  \bibinfo{person}{Pascal Frossard}.} \bibinfo{year}{2017}\natexlab{}.
\newblock \showarticletitle{Universal Adversarial Perturbations}. In
  \bibinfo{booktitle}{\emph{Computer Vision and Pattern Recognition (CVPR),
  2017 IEEE Conference on}}. IEEE, \bibinfo{pages}{86--94}.
\newblock


\bibitem[\protect\citeauthoryear{Moosavi~Dezfooli, Fawzi, and
  Frossard}{Moosavi~Dezfooli et~al\mbox{.}}{2016}]%
        {moosavi2016deepfool}
\bibfield{author}{\bibinfo{person}{Seyed~Mohsen Moosavi~Dezfooli},
  \bibinfo{person}{Alhussein Fawzi}, {and} \bibinfo{person}{Pascal Frossard}.}
  \bibinfo{year}{2016}\natexlab{}.
\newblock \showarticletitle{Deepfool: a simple and accurate method to fool deep
  neural networks}. In \bibinfo{booktitle}{\emph{Proceedings of 2016 IEEE
  Conference on Computer Vision and Pattern Recognition (CVPR)}}.
\newblock


\bibitem[\protect\citeauthoryear{Mopuri, Garg, and Babu}{Mopuri
  et~al\mbox{.}}{2017a}]%
        {mopuri2017fast}
\bibfield{author}{\bibinfo{person}{Konda~Reddy Mopuri}, \bibinfo{person}{Utsav
  Garg}, {and} \bibinfo{person}{R~Venkatesh Babu}.}
  \bibinfo{year}{2017}\natexlab{a}.
\newblock \showarticletitle{Fast Feature Fool: A data independent approach to
  universal adversarial perturbations}.
\newblock \bibinfo{journal}{\emph{arXiv preprint arXiv:1707.05572}}
  (\bibinfo{year}{2017}).
\newblock


\bibitem[\protect\citeauthoryear{Mopuri, Ojha, Garg, and Babu}{Mopuri
  et~al\mbox{.}}{2017b}]%
        {mopuri2017nag}
\bibfield{author}{\bibinfo{person}{Konda~Reddy Mopuri},
  \bibinfo{person}{Utkarsh Ojha}, \bibinfo{person}{Utsav Garg}, {and}
  \bibinfo{person}{R~Venkatesh Babu}.} \bibinfo{year}{2017}\natexlab{b}.
\newblock \showarticletitle{NAG: Network for Adversary Generation}.
\newblock \bibinfo{journal}{\emph{arXiv preprint arXiv:1712.03390}}
  (\bibinfo{year}{2017}).
\newblock


\bibitem[\protect\citeauthoryear{Nair and Hinton}{Nair and Hinton}{2010}]%
        {nair2010rectified}
\bibfield{author}{\bibinfo{person}{Vinod Nair} {and}
  \bibinfo{person}{Geoffrey~E Hinton}.} \bibinfo{year}{2010}\natexlab{}.
\newblock \showarticletitle{Rectified linear units improve restricted boltzmann
  machines}. In \bibinfo{booktitle}{\emph{Proceedings of the 27th international
  conference on machine learning (ICML-10)}}. \bibinfo{pages}{807--814}.
\newblock


\bibitem[\protect\citeauthoryear{Net}{Net}{2016}]%
        {malware2016zdnet}
\bibfield{author}{\bibinfo{person}{ZD Net}.} \bibinfo{year}{ZD
  Net,2016}\natexlab{}.
\newblock \bibinfo{title}{{Surveillance cameras sold on Amazon infected with
  malware}}.
\newblock
  \bibinfo{howpublished}{\url{https://www.zdnet.com/article/amazon-surveillance-cameras-infected-with-malware/}}.
\newblock
\newblock
\shownote{[Online; accessed 30-April-2018].}


\bibitem[\protect\citeauthoryear{Papernot, Carlini, Goodfellow, Feinman,
  Faghri, Matyasko, Hambardzumyan, Juang, Kurakin, Sheatsley,
  et~al\mbox{.}}{Papernot et~al\mbox{.}}{2016a}]%
        {papernot2016cleverhans}
\bibfield{author}{\bibinfo{person}{Nicolas Papernot}, \bibinfo{person}{Nicholas
  Carlini}, \bibinfo{person}{Ian Goodfellow}, \bibinfo{person}{Reuben Feinman},
  \bibinfo{person}{Fartash Faghri}, \bibinfo{person}{Alexander Matyasko},
  \bibinfo{person}{Karen Hambardzumyan}, \bibinfo{person}{Yi-Lin Juang},
  \bibinfo{person}{Alexey Kurakin}, \bibinfo{person}{Ryan Sheatsley},
  {et~al\mbox{.}}} \bibinfo{year}{2016}\natexlab{a}.
\newblock \showarticletitle{cleverhans v2. 0.0: an adversarial machine learning
  library}.
\newblock \bibinfo{journal}{\emph{arXiv preprint arXiv:1610.00768}}
  (\bibinfo{year}{2016}).
\newblock


\bibitem[\protect\citeauthoryear{Papernot, McDaniel, and Goodfellow}{Papernot
  et~al\mbox{.}}{2016b}]%
        {papernot2016transferability}
\bibfield{author}{\bibinfo{person}{Nicolas Papernot}, \bibinfo{person}{Patrick
  McDaniel}, {and} \bibinfo{person}{Ian Goodfellow}.}
  \bibinfo{year}{2016}\natexlab{b}.
\newblock \showarticletitle{Transferability in machine learning: from phenomena
  to black-box attacks using adversarial samples}.
\newblock \bibinfo{journal}{\emph{arXiv preprint arXiv:1605.07277}}
  (\bibinfo{year}{2016}).
\newblock


\bibitem[\protect\citeauthoryear{Radford, Metz, and Chintala}{Radford
  et~al\mbox{.}}{2015}]%
        {radford2015unsupervised}
\bibfield{author}{\bibinfo{person}{Alec Radford}, \bibinfo{person}{Luke Metz},
  {and} \bibinfo{person}{Soumith Chintala}.} \bibinfo{year}{2015}\natexlab{}.
\newblock \showarticletitle{Unsupervised representation learning with deep
  convolutional generative adversarial networks}.
\newblock \bibinfo{journal}{\emph{arXiv preprint arXiv:1511.06434}}
  (\bibinfo{year}{2015}).
\newblock


\bibitem[\protect\citeauthoryear{Sharif, Bhagavatula, Bauer, and Reiter}{Sharif
  et~al\mbox{.}}{2016}]%
        {sharif2016accessorize}
\bibfield{author}{\bibinfo{person}{Mahmood Sharif}, \bibinfo{person}{Sruti
  Bhagavatula}, \bibinfo{person}{Lujo Bauer}, {and} \bibinfo{person}{Michael~K
  Reiter}.} \bibinfo{year}{2016}\natexlab{}.
\newblock \showarticletitle{Accessorize to a crime: Real and stealthy attacks
  on state-of-the-art face recognition}. In
  \bibinfo{booktitle}{\emph{Proceedings of the 2016 ACM SIGSAC Conference on
  Computer and Communications Security}}. ACM, \bibinfo{pages}{1528--1540}.
\newblock


\bibitem[\protect\citeauthoryear{Sharif, Bhagavatula, Bauer, and Reiter}{Sharif
  et~al\mbox{.}}{2017}]%
        {sharif2017adversarial}
\bibfield{author}{\bibinfo{person}{Mahmood Sharif}, \bibinfo{person}{Sruti
  Bhagavatula}, \bibinfo{person}{Lujo Bauer}, {and} \bibinfo{person}{Michael~K
  Reiter}.} \bibinfo{year}{2017}\natexlab{}.
\newblock \showarticletitle{Adversarial Generative Nets: Neural Network Attacks
  on State-of-the-Art Face Recognition}.
\newblock \bibinfo{journal}{\emph{arXiv preprint arXiv:1801.00349}}
  (\bibinfo{year}{2017}).
\newblock


\bibitem[\protect\citeauthoryear{Soomro, Zamir, and Shah}{Soomro
  et~al\mbox{.}}{2012}]%
        {soomro2012ucf101}
\bibfield{author}{\bibinfo{person}{Khurram Soomro},
  \bibinfo{person}{Amir~Roshan Zamir}, {and} \bibinfo{person}{Mubarak Shah}.}
  \bibinfo{year}{2012}\natexlab{}.
\newblock \showarticletitle{UCF101: A dataset of 101 human actions classes from
  videos in the wild}.
\newblock \bibinfo{journal}{\emph{arXiv preprint arXiv:1212.0402}}
  (\bibinfo{year}{2012}).
\newblock


\bibitem[\protect\citeauthoryear{Sultani, Chen, and Shah}{Sultani
  et~al\mbox{.}}{2018}]%
        {sultani2018real}
\bibfield{author}{\bibinfo{person}{Waqas Sultani}, \bibinfo{person}{Chen Chen},
  {and} \bibinfo{person}{Mubarak Shah}.} \bibinfo{year}{2018}\natexlab{}.
\newblock \showarticletitle{Real-world Anomaly Detection in Surveillance
  Videos}.
\newblock \bibinfo{journal}{\emph{arXiv preprint arXiv:1801.04264}}
  (\bibinfo{year}{2018}).
\newblock


\bibitem[\protect\citeauthoryear{Szegedy, Zaremba, Sutskever, Bruna, Erhan,
  Goodfellow, and Fergus}{Szegedy et~al\mbox{.}}{2013}]%
        {szegedy2013intriguing}
\bibfield{author}{\bibinfo{person}{Christian Szegedy},
  \bibinfo{person}{Wojciech Zaremba}, \bibinfo{person}{Ilya Sutskever},
  \bibinfo{person}{Joan Bruna}, \bibinfo{person}{Dumitru Erhan},
  \bibinfo{person}{Ian Goodfellow}, {and} \bibinfo{person}{Rob Fergus}.}
  \bibinfo{year}{2013}\natexlab{}.
\newblock \showarticletitle{Intriguing properties of neural networks}.
\newblock \bibinfo{journal}{\emph{arXiv preprint arXiv:1312.6199}}
  (\bibinfo{year}{2013}).
\newblock


\bibitem[\protect\citeauthoryear{Tensorflow}{Tensorflow}{2016}]%
        {c3dgit}
\bibfield{author}{\bibinfo{person}{C3D Tensorflow}.}
  \bibinfo{year}{2016}\natexlab{}.
\newblock \bibinfo{title}{{C3D Implementation}}.
\newblock
  \bibinfo{howpublished}{\url{https://github.com/hx173149/C3D-tensorflow.git}}.
\newblock
\newblock
\shownote{[Online; accessed 30-April-2018].}


\bibitem[\protect\citeauthoryear{Tram{\`e}r, Kurakin, Papernot, Boneh, and
  McDaniel}{Tram{\`e}r et~al\mbox{.}}{2017}]%
        {tramer2017ensemble}
\bibfield{author}{\bibinfo{person}{Florian Tram{\`e}r}, \bibinfo{person}{Alexey
  Kurakin}, \bibinfo{person}{Nicolas Papernot}, \bibinfo{person}{Dan Boneh},
  {and} \bibinfo{person}{Patrick McDaniel}.} \bibinfo{year}{2017}\natexlab{}.
\newblock \showarticletitle{Ensemble adversarial training: Attacks and
  defenses}.
\newblock \bibinfo{journal}{\emph{arXiv preprint arXiv:1705.07204}}
  (\bibinfo{year}{2017}).
\newblock


\bibitem[\protect\citeauthoryear{Tran, Bourdev, Fergus, Torresani, and
  Paluri}{Tran et~al\mbox{.}}{2015}]%
        {tran2015learning}
\bibfield{author}{\bibinfo{person}{Du Tran}, \bibinfo{person}{Lubomir Bourdev},
  \bibinfo{person}{Rob Fergus}, \bibinfo{person}{Lorenzo Torresani}, {and}
  \bibinfo{person}{Manohar Paluri}.} \bibinfo{year}{2015}\natexlab{}.
\newblock \showarticletitle{Learning spatiotemporal features with 3d
  convolutional networks}. In \bibinfo{booktitle}{\emph{Computer Vision (ICCV),
  2015 IEEE International Conference on}}. IEEE, \bibinfo{pages}{4489--4497}.
\newblock


\bibitem[\protect\citeauthoryear{Tripathi, Mittal, Gangodkar, and
  Kanth}{Tripathi et~al\mbox{.}}{2016}]%
        {tripathi2016real}
\bibfield{author}{\bibinfo{person}{Vikas Tripathi}, \bibinfo{person}{Ankush
  Mittal}, \bibinfo{person}{Durgaprasad Gangodkar}, {and}
  \bibinfo{person}{Vishnu Kanth}.} \bibinfo{year}{2016}\natexlab{}.
\newblock \showarticletitle{Real time security framework for detecting abnormal
  events at ATM installations}.
\newblock \bibinfo{journal}{\emph{Journal of Real-Time Image Processing}}
  (\bibinfo{year}{2016}), \bibinfo{pages}{1--11}.
\newblock


\bibitem[\protect\citeauthoryear{Varol, Laptev, and Schmid}{Varol
  et~al\mbox{.}}{2017}]%
        {varol2017long}
\bibfield{author}{\bibinfo{person}{Gul Varol}, \bibinfo{person}{Ivan Laptev},
  {and} \bibinfo{person}{Cordelia Schmid}.} \bibinfo{year}{2017}\natexlab{}.
\newblock \showarticletitle{Long-term temporal convolutions for action
  recognition}.
\newblock \bibinfo{journal}{\emph{IEEE transactions on pattern analysis and
  machine intelligence}} (\bibinfo{year}{2017}).
\newblock


\bibitem[\protect\citeauthoryear{Vision}{Vision}{[n. d.]a}]%
        {umboocv2017caseTai}
\bibfield{author}{\bibinfo{person}{Umboo~Computer Vision}.} \bibinfo{year}{[n.
  d.]}\natexlab{a}.
\newblock \bibinfo{title}{{Case Study: Elementary Scholl in Taiwai}}.
\newblock
  \bibinfo{howpublished}{\url{https://news.umbocv.com/case-study-taiwan-elementary-school-13fa14cdb167}}.
\newblock


\bibitem[\protect\citeauthoryear{Vision}{Vision}{[n. d.]b}]%
        {umboocv2017caseNCHU}
\bibfield{author}{\bibinfo{person}{Umboo~Computer Vision}.} \bibinfo{year}{[n.
  d.]}\natexlab{b}.
\newblock \bibinfo{title}{{Umbo Customer Case Study NCHU}}.
\newblock
  \bibinfo{howpublished}{\url{https://news.umbocv.com/umbo-customer-case-study-nchu-687356292f43}}.
\newblock


\bibitem[\protect\citeauthoryear{Vision}{Vision}{[n. d.]c}]%
        {umboocv2017caseCBS}
\bibfield{author}{\bibinfo{person}{Umboo~Computer Vision}.} \bibinfo{year}{[n.
  d.]}\natexlab{c}.
\newblock \bibinfo{title}{{Umbo's Smart City Featured on CBS Sacramento}}.
\newblock
  \bibinfo{howpublished}{\url{https://news.umbocv.com/umbos-smart-city-featured-on-cbs-sacramento-26f839415c51}}.
\newblock


\bibitem[\protect\citeauthoryear{Vision}{Vision}{2016}]%
        {umboocv2016case}
\bibfield{author}{\bibinfo{person}{Umboo~Computer Vision}.}
  \bibinfo{year}{2016}\natexlab{}.
\newblock \bibinfo{title}{{Case Studies}}.
\newblock
  \bibinfo{howpublished}{\url{https://news.umbocv.com/case-studies/home}}.
\newblock
\newblock
\shownote{[Online; accessed 30-April-2018].}


\bibitem[\protect\citeauthoryear{Vondrick, Pirsiavash, and Torralba}{Vondrick
  et~al\mbox{.}}{2016}]%
        {vondrick2016generating}
\bibfield{author}{\bibinfo{person}{Carl Vondrick}, \bibinfo{person}{Hamed
  Pirsiavash}, {and} \bibinfo{person}{Antonio Torralba}.}
  \bibinfo{year}{2016}\natexlab{}.
\newblock \showarticletitle{Generating videos with scene dynamics}. In
  \bibinfo{booktitle}{\emph{Advances In Neural Information Processing
  Systems}}. \bibinfo{pages}{613--621}.
\newblock


\bibitem[\protect\citeauthoryear{Xu, Evans, and Qi}{Xu et~al\mbox{.}}{2017a}]%
        {xu2017feature2}
\bibfield{author}{\bibinfo{person}{Weilin Xu}, \bibinfo{person}{David Evans},
  {and} \bibinfo{person}{Yanjun Qi}.} \bibinfo{year}{2017}\natexlab{a}.
\newblock \showarticletitle{Feature squeezing: Detecting adversarial examples
  in deep neural networks}.
\newblock \bibinfo{journal}{\emph{arXiv preprint arXiv:1704.01155}}
  (\bibinfo{year}{2017}).
\newblock


\bibitem[\protect\citeauthoryear{Xu, Evans, and Qi}{Xu et~al\mbox{.}}{2017b}]%
        {xu2017feature1}
\bibfield{author}{\bibinfo{person}{Weilin Xu}, \bibinfo{person}{David Evans},
  {and} \bibinfo{person}{Yanjun Qi}.} \bibinfo{year}{2017}\natexlab{b}.
\newblock \showarticletitle{Feature squeezing mitigates and detects
  carlini/wagner adversarial examples}.
\newblock \bibinfo{journal}{\emph{arXiv preprint arXiv:1705.10686}}
  (\bibinfo{year}{2017}).
\newblock


\bibitem[\protect\citeauthoryear{Zeiler}{Zeiler}{2012}]%
        {zeiler2012adadelta}
\bibfield{author}{\bibinfo{person}{Matthew~D Zeiler}.}
  \bibinfo{year}{2012}\natexlab{}.
\newblock \showarticletitle{ADADELTA: an adaptive learning rate method}.
\newblock \bibinfo{journal}{\emph{arXiv preprint arXiv:1212.5701}}
  (\bibinfo{year}{2012}).
\newblock


\bibitem[\protect\citeauthoryear{Zhu, Kr{\"a}henb{\"u}hl, Shechtman, and
  Efros}{Zhu et~al\mbox{.}}{2016}]%
        {zhu2016generative}
\bibfield{author}{\bibinfo{person}{Jun-Yan Zhu}, \bibinfo{person}{Philipp
  Kr{\"a}henb{\"u}hl}, \bibinfo{person}{Eli Shechtman}, {and}
  \bibinfo{person}{Alexei~A Efros}.} \bibinfo{year}{2016}\natexlab{}.
\newblock \showarticletitle{Generative visual manipulation on the natural image
  manifold}. In \bibinfo{booktitle}{\emph{European Conference on Computer
  Vision}}. Springer, \bibinfo{pages}{597--613}.
\newblock


\end{thebibliography}

\end{document}